\documentclass[aic, crcready]{iosart2x}

\usepackage[absolute]{textpos}
\usepackage{latexsym}
\usepackage{amssymb}
\usepackage{amsmath}
\usepackage{amsthm}
\usepackage{booktabs}
\usepackage{enumitem}
\usepackage{graphicx}
\usepackage{color}

\usepackage{array}
\usepackage{multirow}
\usepackage{soul}

\usepackage{pifont}  

\usepackage{makecell}  

\usepackage{xcolor}

\newcommand{\CellWithForceBreak}[2][c]{\begin{tabular}[#1]{@{}c@{}}#2\end{tabular}}

\newcommand{\redcross}{\textcolor{red!80!black}{\ding{55}}}

\newlength{\symbolwidth}
\pubyear{0000}
\volume{0}
\firstpage{1}
\lastpage{1}

\begin{document}

\begin{textblock*}{20cm}(4cm, 1cm) 
    \noindent \textsf{\small Preprint version. Accepted for publication in \textit{The European Journal on Artificial Intelligence} (2026).}
\end{textblock*}

\begin{frontmatter}

\title{Interpretability of the Intent Detection Problem: A New Approach}
\runtitle{Interpretability of the Intent Detection Problem: A New Approach}


\begin{aug}
\author[A]{\inits{E.}\fnms{Eduardo} \snm{Sanchez-Karhunen}\ead[label=e1]{fesanchez@us.es}%
\thanks{Corresponding author. \printead{e1}.}}
\author[A]{\inits{N.N.}\fnms{Jose F.} \snm{Quesada-Moreno}\ead[label=e2]{jquesada@us.es}}
\author[A]{\inits{N.-N.}\fnms{Miguel A.} \snm{Guti\'errez-Naranjo}\ead[label=e3]{magutier@us.es}}
\address[A]{Departamento de Ciencias de la Computación e Inteligencia Artificial, \orgname{Universidad de Sevilla, Avda. Reina Mercedes s/n, 41012, Sevilla, Andalucía,} \cny{Spain}.\printead[presep={\\}]{e1,e2,e3}}
\end{aug}

\begin{abstract}
\noindent Intent detection, a fundamental text classification task, aims to identify and label the semantics of user queries, playing a vital role in numerous business applications. Despite the dominance of deep learning techniques in this field, the internal mechanisms enabling Recurrent Neural Networks (RNNs) to solve intent detection tasks are poorly understood. In this work, we apply dynamical systems theory to analyze how RNN architectures address this problem, using both the balanced SNIPS and the imbalanced ATIS datasets. By interpreting sentences as trajectories in the hidden state space, we first show that on the balanced SNIPS dataset, the network learns an ideal solution: the state space, constrained to a low-dimensional manifold, is partitioned into distinct clusters corresponding to each intent.  The application of this framework to the imbalanced ATIS dataset then reveals how this ideal geometric solution is distorted by class imbalance, causing the clusters for low-frequency intents to degrade. Our framework decouples geometric separation from readout alignment, providing a novel, mechanistic explanation for real world performance disparities. These findings provide new insights into RNN dynamics, offering a geometric interpretation of how dataset properties directly shape a network's computational solution.
\end{abstract}


\end{frontmatter}

\section{Introduction}
\subsection[RNNs and the Interpretability Challenge]{RNNs and the Interpretability Challenge\footnote{This paper is an extended version of the work presented at 27th European Conference on Artificial Intelligence \cite{sanchez-karhunen_interpretation_2024}}}
Modern recurrent neural networks (RNNs) are widely used to tackle problems involving sequential data. These networks have demonstrated strong performance in various natural language processing (NLP) tasks, such as sentiment analysis \cite{liu_sentiment_2015}, intent detection and slot filling \cite{hakkani-tur_multi-domain_2016}, and machine translation \cite{sutskever_sequence_2014}. However, despite their widespread success, the exact nature of the internal mechanisms by which RNNs solve specific tasks remains an open question. This lack of understanding is partly due to the nonlinear nature of RNNs and the high-dimensionality of their hidden layers, which together obscure the computational processes underlying their behavior. Moreover, the trend in practical applications leans towards increasingly complex architectures \cite{bahdanau_neural_2015, sutskever_sequence_2014}, making it even more challenging to understand what is happening inside the nets. The integration of RNNs into applications with significant societal impact, such as healthcare, legal systems, and autonomous decision-making, has made the demand for interpretability more pressing than ever. Understanding how these models detect patterns, solve problems, and make decisions is no longer merely a theoretical concern but a practical necessity. Enhancing the interpretability of neural networks decision-making is crucial to ensure their robustness, fairness, and accountability in real-world applications \cite{hamon_robustness_2020}. Addressing this challenge is key to bridge the gap between the impressive capabilities of RNNs and the trust required for their deployment in high-stakes environments.

\subsection{Analysis Approaches for Understanding RNNs}
Understanding the behavior of RNNs has been a persistent challenge in machine learning research. Early studies focused on visualizing the activity of specific network components, such as memory gates, during NLP tasks \cite{karpathy_visualizing_2016, strobelt_lstmvis_2018}. Although these unit-level analyses offer localized functions, they often fail to provide a comprehensive interpretation of the network's overall behavior. The inherent feedback connections between RNN neurons allow these networks to be viewed as nonlinear dynamical systems \cite{martelli_introduction_1999}, allowing the application of well-established tools from dynamical systems theory \cite{strogatz_nonlinear_2015}. Based on this perspective, several studies have derived analytic expressions for aspects of network dynamics such as bifurcations in the parameter space of small networks and convergence properties \cite{haschke_input_2005, yi_convergence_2004}. These efforts have enhanced our understanding of the mathematical underpinnings of RNN behavior but remain limited in their ability to explain task-specific computations in large, real-world networks. Recently, a new reverse engineering paradigm has emerged to analyze RNNs at a higher level of abstraction. Instead of focusing on the microdetails of individual neurons or gates, this approach examines the state space of trained RNNs. Fixed points are identified and the dynamics of the system is linearized around them, revealing a key computational mechanism embedded within the network \cite{sussillo_opening_2013}. This perspective has yielded significant insights, particularly in the context of text classification tasks, where state space analyses have demonstrated promising results \cite{aitken_geometry_2021,maheswaranathan_reverse_2019}. A recurring theme in these works is the discovery that trained RNNs often converge to highly interpretable, low-dimensional representations associated with attractors in the state space. The geometry and dimensionality of these attractors manifolds are intricately linked to the structure of the dataset and the nature of the task being solved. In summary, this attractor-based view provides a powerful framework for understanding how RNNs encode information and implement computations.  

\subsection{A Comparison with Other Interpretability Paradigms}
Our analysis adopts a dynamical systems framework to seek a mechanistic understanding of RNN computation, offering insights distinct from other valuable interpretability paradigms. Contrasting our global, dynamic approach with other common methods clarifies its unique contributions. \textbf{Input-attribution methods} aim to attribute a network's output back to specific input tokens. Attention mechanisms \citep{bahdanau_neural_2015}, assign relevance scores to input tokens, highlighting what parts of an input sequence the model focuses on when generating an output. Similarly, gradient-based methods like saliency maps \citep{simonyan_salience_2013} compute the output's sensitivity to small changes in the input, identifying influential features. While these methods excel at answering which input features are most influential, they largely treat the recurrent core as a black box. They offer limited insight into the internal computational processes that transform information into a final classification. \textbf{Local surrogate methods:} techniques like LIME \citep{ribeiro_why_2016} and SHAP \citep{lundberg_unified_2017} operate on a different principle. They explain an individual prediction by creating a simpler, interpretable surrogate model that is faithful to the complex model's behavior in a localized region around the specific input. Their power lies in this instance-specific explanation, but their scope is inherently local and cannot be used to describe the global structure of the state space or the general principles that govern the network's behavior across all possible inputs. 

In contrast, our dynamical systems approach provides a  different and complementary level of insight. 
\begin{itemize}
 \item \textbf{Revealing dynamic processes:} Unlike static attribution maps, this framework visualizes computation as a dynamic process. It allows us to model sentences as trajectories within the hidden state space, revealing how evidence is accumulated token-by-token, rather than just identifying which tokens were most influential.   
\item \textbf{Characterizing global geometry}: Where other methods are local, our approach characterizes the global geometry of the task learned by the network. This enables the analysis of holistic properties such as the intrinsic dimensionality of the problem and the organization of the decision space into meaningful regions.
\item \textbf{Uncovering mechanistic principles}: Most critically, this perspective moves from correlation to mechanism. It seeks to explain why the network functions as it does, framing classification as a process governed by the stable and unstable dynamics of a fixed-point topology. Therefore, our approach complements other methods by providing a unique lens into the internal, dynamic, and geometric nature of how RNNs solve tasks.
\end{itemize}
\subsection{Intent Detection: A Case Study Domain}
Within the field of NLP, intent detection is a horizontal foundational operation, providing instrumental support for a wide range of applications. Broadly defined, intent detection is an NLP task aimed at recognizing and classifying the underlying purpose or operational goal (a.k.a. intention) expressed in a user's utterance. It serves as a critical component in the functional architecture of language understanding systems \cite{qin2021survey}, addressing the challenge of mapping a large and diverse set of linguistic expressions onto a predefined set of semantic intentions. This challenge is complex, requiring the operation of multiple linguistic levels simultaneously. From diverse lexical realizations and syntactic structures, to semantic ambiguities and pragmatic contexts, intent detection must integrate diverse dimensions of language processing to deliver accurate results. Despite progress in academic research and industrial applications, intent detection remains a theoretical and practical challenge. Different lines of interest include: detecting multiple intents within a single utterance \cite{kim_et_al_2017}, integrating intent detection with entity recognition \cite{abro_et_al_2022, weld_et_al_2022, song_et_al_2022}, managing out-of-domain intents \cite{capuano_2021, lang_et_al_2023}, and developing robust models that support explainability and transparency in classification decisions \cite{Zhuang_Cheng_Zou_2024}. These challenges, combined with the intrinsic characteristics of intent detection: a) operation on a low-dimensional semantic space, despite the high-dimensionality of the architectures involved, and b) a topologically inspired convergence towards different semantic kernels, make it an ideal domain for analysis using dynamical systems theory. This work leverages a combined mathematical, computational, and linguistic framework rooted in dynamical systems theory to propose a novel approach to understanding the nature of the intent detection process. This framework also constitutes the starting point for tackling some of the challenges indicated.

\subsection{Our Contributions and Paper Organization}
Our key contribution is the pioneering study of the state-space dynamics of trained RNNs applied to the SNIPS and ATIS intent detection problems. We show that the state space can be characterized as a low-dimensional manifold whose intrinsic dimensionality is related to the size of the embedding layer and the number of neurons in the hidden layer. We show that input sentences traverse discrete trajectories through the state space, progressing from initial states toward specific outer regions. A crucial finding is the identification of distant regions within the state space, where these trajectories terminate. These peripheral areas are aligned with the directions defined by the rows of the readout matrix, allowing for the generation of predictions. In addition, we uncover the fixed-point topology underlying the network dynamics. Unlike other tasks, such as sentiment analysis or document classification, we find that RNNs trained for intent detection exhibit an unexpected fixed-point structure \cite{aitken_geometry_2021}.  The number and nature of attractors, saddle points and other critical points in the state space are shown to depend on network parameters and the type of RNN cell (e.g. LSTM or GRU). We first established this geometric framework on the balanced SNIPS dataset, and then use it as a diagnostic tool to test its generalizability on the complex, imbalanced ATIS dataset. This analysis reveals how the ideal geometric solution is distorted by class imbalance on low-frequency intents. We introduce a novel diagnostic framework that decouples geometric separation from readout alignment, allowing us to identify four distinct, mechanistic patterns that explain real-world performance disparities.

The rest of the paper is organized as follows: Section~\ref{sec:intent_detection_problem} introduces the problem of intent detection, discussing various aspects and current lines of research in this field, as well as how the dynamical systems approach fits into this context. Section~\ref{sec:dynamical_systems} explores how RNNs can be interpreted as nonlinear dynamical systems. Section~\ref{sec:objectives} outlines the specific objectives to be addressed through the experiments. Section~\ref{sec:datasets} describes the datasets selected for this study. Section~\ref{sec:experimental_setup} details the experimental setup and methodology, while Section~\ref{sec:results} provides an in-depth analysis of the results obtained. Finally, Section 8 presents the conclusions and suggest potential directions for future research.

\section{Intent Detection: Domain Characterization and Research Challenges}\label{sec:intent_detection_problem}
Intent detection is a cornerstone in several areas of NLP, playing a vital role in numerous industrial applications, particularly in conversational systems, virtual assistants, and question-answering environments. At its core, intent detection addresses one of the most fundamental syntactic-semantic challenges in natural language understanding: to collaborate effectively in conversational interactions, a conversational agent must accurately identify and unambiguously classify the user's intent. This task depends on various linguistic components and levels. From lexical and morphological units to syntactic, semantic, and even pragmatic and acoustic nuances. In particular, even non-verbal cues, such as silence or pauses, can convey relevant intentional meanings.

\subsection{Evolution of Intent Detection Approaches}
The field of intention recognition has evolved significantly since its inception. Three main paradigms can be distinguished, starting with simple dictionary-based and rule-based methods \cite{Niimi_et_al_2001}. A second major approach focuses on statistically based classification methods \cite{Jansen_et_al_2007}. In the current landscape, intent detection is experiencing a remarkable moment of dynamism, adopting the most advanced techniques in deep learning and natural language processing. Different neural network architectures, from RNNs to transformer-based models, have been applied, including proposals based on models such as BERT \cite{abro_et_al_2022, sevini_et_al_2019, WU2024127054}.

\subsection{Current Research Challenges}
Despite significant progress, intent detection remains a challenging task due to the complexity of human interactions. Given the foundational nature of intent detection, as a core component of different language technology systems, research must address different challenges. 
\begin{itemize}
\item {\bf Multi-intent detection}: User utterance in conversational interactions often includes multiple intents simultaneously \cite{kim_et_al_2017}. For example, in the sentence: {\it ‘Hello good morning, I would like to cancel the medical appointment I had and ask for a new one’} the user expresses three different intents: greeting, cancelation and rescheduling. Moreover, temporal dependencies between intents can arise, as in: {\it ‘I want to make a transfer but first I would like to know my balance’}. In this case, the execution of the second intent depends on the completion of the first. In addition, operational or interpretative dependencies may exist, as in: {\it ‘Turn off all the lights that are on and turn on all the lights that are off’}. In this case, determining which lights are affected by the second intent must be resolved before executing the action associated with the first intent. 

\item {\bf Integration of intent detection and entity recognition}: The tasks of intent detection and entity recognition are deeply interconnected but have traditionally been treated as separate problems within the language understanding pipeline \cite{abro_et_al_2022, weld_et_al_2022, song_et_al_2022}. Entities may be linked to different intents according to semantic or pragmatic relationships. Hence, a correct identification of entities may depend to a large extent on the recognition of intents. For example, in the sentence {\it ‘I would like to book the flight departing at 9’}, the value {\it ‘9’} should be classified as a time, while in {\it ‘I would like to order 9’}, the same value {\it ‘9’} represents a quantity. Furthermore, entity recognition can clarify ambiguous intents, as in {\it ‘From Paris to New York next Monday’}.
\item {\bf Out-of-domain intent management}: Effective handling of out-of-domain intents is essential for real-world applications, particularly in domains like banking, public administration, and customer service \cite{capuano_2021, lang_et_al_2023}. Users often express intents that fall outside the operational scope of a system, posing a challenge for intent detection. This complexity is amplified by a high level of overlap in the way intents can be expressed. For example, a citizen may request the renewal of a document from the wrong administration, but linguistically, the request resembles a legitimate query. Detecting and accurately classifying these cases remains a significant challenge. 
\item {\bf Model explainability}: In many critical applications, it is not sufficient for a conversational system to simply classify the intent but also to provide a clear and understandable explanation associated with the decision \cite{Zhuang_Cheng_Zou_2024}. Moreover, explainability contributes to ensure the robustness and reliability of conversational systems and the identification of potential biases.
\end{itemize}

\subsection{Suitability for Dynamical Systems Analysis}
The analysis of the intention detection task, together with its current research directions and the associated methodological and technological challenges, suggests that it is a highly suitable domain for the application of dynamical systems analysis. As we show in this paper, intent detection models typically operate on low-dimensional manifolds, even though they are implemented using high-dimensional neural architectures. This feature reflects the intrinsic nature of intent detection, which at a high level of abstraction can be described as a mapping operation from complex linguistic expressions to a relatively small catalog of semantic intents. The objective is to maintain strong robustness and flexibility across with respect to variability of expressions at all linguistic levels, including lexical-morphological, syntactic, and semantic variability. On the other hand, modeling based on specific fixed-point topologies with attractors and saddle points aligns well with the inherent structure of intent classification problems. The targets in the intent catalog can ultimately be interpreted as convergence points of the entire set of linguistic variants that express or represent the same meaning (semantic-level cataloging) or are associated with the execution of the same operation (pragmatic-level cataloging). This mathematical and computational framework provides a suitable approach for addressing one of the key characteristics of human language, such as ambiguity and the transition between different interpretations. 

A key intuition underlying this paper is that the process of understanding, which underpins intent classification, can be viewed as a trajectory in a multidimensional space defined by word embeddings as they are processed and transformed by the network in the hidden layer for each token of an input expression. Analyzing these trajectories through space offers valuable insights into the comprehension process, which can be seen as a nonlinear accumulation of information received from the input signal in sequential nature. Within this framework, attractor points can be interpreted as the centers of regions that represent the intents defined in the catalog of the working domain. This approach integrates motivations of a computational nature (rooted in deep learning models with architectures ranging from RNNs to Transformers), mathematical (through the use of a topological model that enables operations such as comparison, trajectory analysis, proximity assessment, sudden changes in direction and speed), and linguistic (addressing issues like ambiguity, synonymy, polysemy, and dependencies between linguistic levels). This work also offers a starting point for tackling the challenges in the field of intent detection. For instance, the state space topology provides a new perspective for addressing multi-intent detection and out-of-domain intent identification. Additionally, the low-dimensional manifold structure serves as a foundation for analyzing the interaction between intent detection and entity recognition within a unified semantic space. Crucially, this framework's ability to provide a mathematical foundation for understanding the internal workings of intent detection systems enhances their explanatory power, addressing the limitations of black-box models and promoting greater transparency.  

\section{Background}\label{sec:dynamical_systems}
\subsection{Recurrent Neural Networks Computations}
\begin{figure}[h]
\centering
\includegraphics[width=13cm]{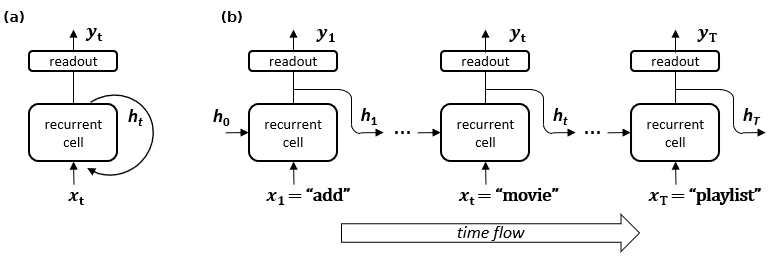}
\caption{\textbf{(a)} Folded representation of a Recurrent Neural Network (RNN), highlighting the recurrent connection within the architecture. \textbf{(b)} Unfolded representation of an RNN, explicitly showing the flow of time. At each time step $t$, the RNN processes an input token $\mathbf{x}_t$ (e.g. words in a sentence like "add", "movie", or "playlist"), updates its hidden state $\mathbf{h}_t$ based on the previous state $\mathbf{h}_{t-1}$, and generates an output $\mathbf{y}_t$. The initial hidden state $\mathbf{h}_0$ represents the starting point of the RNN's before processing any input tokens.}
\label{fig:folded_unfolded}
\end{figure}

Feedforward networks (FFNs) analyze inputs under the assumption that the order in which samples are processed does not carry meaningful information. However, in many real-world scenarios, the order of the components, or their temporal or spatial relationships, is crucial. For example, NLP problems are studied as token sequences, and weather forecasting relies on time series data \cite{han_using_2021}. In such cases, a mechanism for retaining and learning temporal dependencies is required. To address this limitation, FNNs can be enhanced with feedback connections, as illustrated in Figure~\ref{fig:folded_unfolded}. This enhancement gives rise to the architecture known as recurrent neural networks (RNNs) \cite{goodfellow_deep_2016}. 

In general, the computations performed by RNNs can be summarized using the following pair of equations in difference:
\begin{align}
    \hspace{17 em}
    \mathbf{h}_t &= \mathbf{F}(\mathbf{h}_{t-1},\mathbf{x}_t) \label{eq_recurrence}\\
    \mathbf{y}_t &= \mathbf{W}\mathbf{h}_t + \mathbf{b} \label{eq_readout} 
\end{align}

where $t$ represents an integer index (commonly interpreted as time), $\mathbf{h}_t \in \mathbb{R}^n$ is the $n$-dimensional \textit{hidden state} of the network, and $\mathbf{x}_t \in \mathbb{R}^m$ is the $m$-dimensional external input at step $t$. The nonlinear update function $\mathbf{F}$ defines how the hidden state evolves, depending on the specific recurrent cell architecture used (e.g. vanilla RNN, LSTM, GRU). Given an input sequence $\mathbf{x}_1, \dots, \mathbf{x}_T$, the network updates its hidden state $\mathbf{h}_t$ at each step $t$ based on the previous hidden state $\mathbf{h}_{t-1}$ and the current input $\mathbf{x}_t$. The predictions $\mathbf{y}_t$ are computed by passing the hidden states through a linear \textit{readout layer}, where $\mathbf{W}$ is a $n \times n$ readout weight matrix and $\mathbf{b}$ is a bias vector. Each row $\mathbf{r}_i$ of $\mathbf{W}$ is referred to as a readout vector. In classification tasks, the output $\mathbf{y}_t$ consists of $N$ logits, one for each class label. The way these outputs are generated depends on the specific problem. In the so-called many-to-many situations (e.g. named entity recognition), the entire sequence of hidden states $\mathbf{h}_1, \dots, \mathbf{h}_T$ is projected to produce a sequence of predictions $\mathbf{y}_1, \dots, \mathbf{y}_T$. In contrast, in many-to-one contexts (e.g. intent detection or sentiment analysis), only the last hidden state $\mathbf{h}_T$ is used to generate a single prediction $\mathbf{y}_T$.

\subsection{Fixed points}
Systems governed by difference equations as in Equation~\ref{eq_recurrence} are called \textit{discrete-time dynamical systems}, with their state represented by $\mathbf{h}_t$ at time $t$. In the context of RNNs, the update function $\mathbf{F}$ is inherently nonlinear, which classifies these systems as nonlinear discrete-time dynamical systems (NLDS) designed and tuned to perform specific tasks. Therefore, they can be analyzed using a wide variety of tools from the dynamical system theory. The vector state $\mathbf{h}_t \in \mathbb{R}^n$ can be visualized in a $n$-dimensional space known as the \textit{state space} or the \textit{phase space} of the system, where each axis corresponds to a component of $\mathbf{h}_t$. For a given initial state, the system evolves over time according to $\mathbf{F}$, producing a sequence of states that form a \textit{trajectory} or \textit{orbit} in the state space. These trajectories can exhibit highly diverse qualitative behaviors depending on the region of the phase space. As a result,  a common approach is to partition the state space into regions and study the properties and interactions of these areas separately. A natural starting point for this analysis is the study of fixed points. 

A \textit{fixed point} or \textit{equilibrium point}, denoted as $\mathbf{h}^*$, is defined as a zero-motion state (or an invariant point). By definition, if the system reaches $\mathbf{h}^*$, it will remain in it. In practical scenarios, noise or external perturbations may shift the system from a fixed point. Therefore, it is important to study the behavior of the system in the neighborhood of $\mathbf{h}^*$. If the system converges back to the fixed point after a small perturbation, $\mathbf{h}^*$ is classified as a \textit{stable} fixed point. In contrast, if the system diverges from the fixed point, it is called \textit{unstable}. An additional type of fixed point, known as a \textit{saddle point} exhibits a mixed behavior. For certain trajectories, a saddle point behaves like a stable equilibrium, while for others, it behaves as an unstable point.

\subsection{Linearization}
Fixed points have a key property; the Hartman-Grobman theorem \cite{hartman_ordinary_2002} states that the behavior of an NLDS near a fixed point $\mathbf{h}^*$ is qualitatively the same as the behavior of its linearization, provided that the fixed point is hyperbolic.\footnote{A hyperbolic fixed point is one where none of the eigenvalues of the Jacobian have a magnitude equal to 1. For further details, see \cite{hartman_ordinary_2002}.} This equivalence means that the complex dynamics of an NLDS can be approximated locally by a linear system, greatly simplifying the analysis. The analysis of NLDS around fixed points is therefore performed in two steps: a) identifying the fixed points of the system, and b) analyzing the linearized behavior near these points. Linearization is obtained by calculating the \textit{Jacobian matrix} of the system $\mathbf{JF}$ around the fixed point. This Jacobian provides a first-order approximation of the dynamics of the system in the vicinity of $\mathbf{h}^*$. Linearized system analysis involves decomposing the Jacobian matrix as $\mathbf{J = R}$ $\Lambda$ $\mathbf{L}$, where $\Lambda$ is a diagonal matrix that contains the eigenvalues $\lambda_i$ of the $n$-dimensional system, and $\mathbf{L} = \mathbf{R}^{-1}$. Each eigenvalue $\lambda_i$ has an associated eigenvector $\mathbf{r}_i$, which corresponds to a row of the matrix $\mathbf{R}$. 

From this decomposition, the state of a $n$-dimensional linear dynamical system can be expressed as the linear composition of $n$ independent one-dimensional exponential dynamics, often referred to as \textit{modes} or patterns of activity. Each mode evolves along a direction in the state space defined by the eigenvector $\mathbf{v}_i$, which represents an invariant line in the phase space. Consequently, the motion near $\mathbf{h}^*$ can be understood as the linear combination of these $n$ one-dimensional systems. The dynamic of a mode along the direction $\mathbf{v}_i$ is controlled by its associated eigenvalue $\lambda_i$, and the evolution of the system along that direction is given by $\lambda_i^tb_i$, where $\mathbf{b}_i$ is the initial amplitude of the mode. The magnitude of $|\lambda_i|$ determines the stability of the motion along $\mathbf{v}_i$. If $|\lambda_i| > 1$, the component of $\mathbf{h}_t$ along $\mathbf{v}_i$ grows exponentially, indicating an unstable direction. In contrast, if $|\lambda_i| < 1$, the component shrinks exponentially, which indicates a stable direction. The overall stability of the fixed point $\mathbf{h}^*$ is determined by the \textit{spectral radius} of the Jacobian matrix, which is the largest magnitude among the eigenvalues of $\mathbf{JF}(\mathbf{h}^*)$. Specifically, if all the eigenvalues $|\lambda_i|$ are within the unit circle, the fixed point $\mathbf{h}^*$ is stable. If at least one eigenvalue satisfies $|\lambda_i| > 1$, the fixed point is unstable. For saddle points, some eigenvalues lie within the unit circle, while others lie outside, resulting in both stable and unstable directions. Finally, if all eigenvalues $\lambda_i$ lie outside the unit circle, $\mathbf{h}^*$ is a totally unstable equilibrium point.

\subsection{Basins of Attraction and Saddle Points}
In many systems without external input, the dynamics naturally evolves toward specific regions of the state space. These converging points or regions are known as \textit{attractors}. Among the attractors, the simplest type is the stable fixed point. The region of the phase space (i.e. the set of all initial states) from which the system evolves toward a particular attractor is called the \textit{basin of attraction} of an attractor. Any initial condition within this region will eventually lead the system to the attractor through the iterative dynamics of the system. The state space is typically partitioned into basins of attraction, each associated with a specific attractor. Saddle points play a crucial role in governing the boundaries and interactions between these basins. Typically, a saddle point will have a dominant set of stable modes (or manifolds) with only a small number of unstable modes. This configuration makes saddle points critical for state space management, as they influence how trajectories transition between basins of attraction. For example, a region of state space may be funneled through the stable modes of a saddle point, only to be directed toward different attractors by its unstable modes. In this way, saddle points act as gateways or intermediaries that connect different basins of attraction. As a result of this interaction, the stable manifold of a saddle point often forms the boundary between the basins of attraction \cite{ceni_interpreting_2020}. The complexity of a saddle point can be quantified by its \textit{index}, defined as the number of unstable manifolds (or directions) associated with the fixed point. A saddle point with a higher index has a greater number of directions in which the trajectories diverge, which can lead to more intricate dynamics and transitions between basins.

\subsection{Reverse Engineering RNNs of Classification Tasks}
Recurrent Neural Networks, as nonlinear discrete-time dynamical systems, can be analyzed using tools from dynamical system theory. Modern RNN architectures are made up of hidden layers with hundreds of neurons, resulting in high-dimensional hidden states. Traditional dynamical system analysis often treats individual neurons as system parameters, but this high dimensionality poses significant challenges for standard state-space analysis. A recent line of research adopts a higher-level perspective to study the computational mechanisms learned by RNNs \cite{sussillo_opening_2013}. These reverse engineering techniques aim to uncover how trained RNNs implement specific tasks by analyzing the geometry and dynamics of their state space. This approach focuses on key dynamical features, such as fixed points, their linearized dynamics, and the interactions between equilibrium points, to infer the behavior of the network.

For example, tasks such as binary sentiment analysis and general text classification exhibit a common underlying dynamical mechanism \cite{aitken_geometry_2021}. In such tasks, the hidden state trajectories of RNNs largely lie in a low-dimensional subspace of the full state space, despite the high dimensionality of the hidden states. Within this subspace lies an attractor manifold, which serves as a repository for accumulating evidence for each class as the network processes tokens sequentially. The exact dimensionality and geometry of this attractor manifold depend on the structure of the dataset and the complexity of the task. For binary sentiment classification, hidden states typically evolve along a line of stable fixed points, reflecting the network's progression as it processes input text \cite{maheswaranathan_reverse_2019}. More generally, for categorical classification tasks with $N$ classes, the attractors form an $(N-1)$-dimensional simplex that captures the scalar quantities that the network must maintain to accurately classify the input \cite{aitken_geometry_2021}.

\section{Objectives}\label{sec:objectives}
RNNs are nonlinear dynamical systems with high-dimensional state-space structures. These structures can be analyzed by examining fixed points, attractor manifolds, and trajectories within the state space. Recent advances in reverse engineering techniques have provided valuable insight into how RNNs implement task-specific computations. These studies suggest that RNNs encode evidence in low-dimensional manifolds, using an integrative mechanism to track and accumulate information over time to facilitate accurate classification. Although significant progress has been made in understanding RNN dynamics for tasks such as binary sentiment classification and generic text classification, these techniques have not yet been systematically applied to intent detection. Intent detection presents unique challenges due to its diverse and semantically rich set of intent classes. As a result, it remains unclear how state-space dynamics manifest in intent detection tasks and how architectural parameters affect the underlying trajectories and manifold geometry. The primary objective of this study is to investigate how RNN architectures encode intent detection tasks in their state space. Specifically, our goal is to analyze the structure of the state space, determining its intrinsic dimensionality, and comparing it with previous findings for generic categorical text classification tasks.  We also examine the behavior of hidden state trajectories as the RNN processes input sequences, uncovering the internal mechanism that lead to accurate predictions. To address these objectives, we performed experiments using the SNIPS and ATIS datasets and analyzed the state-space dynamics of various RNN architectures.

\section{Datasets}\label{sec:datasets}
Several benchmark datasets have been widely used to evaluate the performance of intent detection models. Among them, three prominent datasets are SNIPS, ATIS, and MASSIVE, each with distinct characteristics and challenges. The SNIPS dataset \cite{coucke_snips_2018} developed for English voice assistant systems consists of 7 balanced intents. This simplicity, combined with the absence of significant class imbalance, makes SNIPS an ideal choice for analyzing the state space dynamics of intent detection models without introducing side effects due to imbalance or an excessive number of classes. In contrast, the ATIS (Airline Travel Information System) dataset \cite{hemphill_atis_1990} contains real customer conversations about flight information. Although ATIS includes 26 intentions, almost 74\% of its samples belong to a single intent, making it challenging to analyze model performance fairly across all classes. Furthermore, some utterances in ATIS are annotated with multiple intents. The recently introduced MASSIVE dataset \cite{fitzgerald_massive_2022} is a multilingual localization of the SLURP dataset \cite{bastianelli_slurp_2020}. It spans 51 languages and includes 60 intents in 18 domains. However, the MASSIVE dataset exhibits both a large number of intents and a degree of class imbalance, with some intents having very few samples. This makes MASSIVE more suitable for multilingual and large-scale evaluations. 

Given our focus on understanding the state space dynamics of RNNs in intent detection tasks, we select two complementary datasets for our experiments: SNIPS and ATIS. SNIPS serves as our baseline. Its reduced number of intents, balanced class distribution, and manageable complexity allow for a controlled investigation of the underlying computational mechanisms. ATIS serves as our generalizability test. Its significant class imbalance and larger number of intents provides a challenging, real-world scenario to validate our framework. Table~\ref{tab:datasets_comparison} compares the key characteristics of these datasets, highlighting differences in class distribution, imbalance, and scale.

\begin{table}[h]
\caption{Comparison of intent detection datasets, summarizing key characteristics, including the number of languages, utterances per language, domains, intents, slots, and class imbalance measures (number and percentage of samples in the largest and smallest intent classes for each dataset).}
\centering
\begin{tabular}{lccccccccc} 
\toprule
\CellWithForceBreak{Dataset \\ Name} & \CellWithForceBreak{Languages \\ \#} & \CellWithForceBreak{Utterances \\ per Language} & \CellWithForceBreak{Domains \\ \#} & \CellWithForceBreak{Intents \\ \#} & \CellWithForceBreak{Slots \\ \#}& \CellWithForceBreak{Samples in\\ Largest Intent} & \CellWithForceBreak{Largest \\ Intent (\%)} & \CellWithForceBreak{Samples in \\ Smallest Intent} & \CellWithForceBreak{Smallest \\ Intent (\%)} \\
\midrule
SNIPS & 1 & 14484 & - & 7 & 53 & 2100 & 14.5 & 2042 & 14.1 \\
ATIS &  1 & 5871 & 1 & 26 & 129 &  4298 & 73.7 & 1 & 0.02\\
MASSIVE & 51 & 19521 & 18 & 60 & 55 & 1190 & 6.9 & 6 & 0.04\\
\bottomrule
\end{tabular}
\label{tab:datasets_comparison}
\end{table}

\section{Experiments Setup}\label{sec:experimental_setup}
Our analysis is structured into four steps: a) train different RNN architectures to solve the intent detection task on the SNIPS and ATIS datasets; b) extract and visualize the state space learned by the RNNs; c) analyze the manifold structure in which the state space is embedded; and d) investigate the fixed-point structure underlying the RNNs.

We implemented and trained the RNN models using TensorFlow 2 \cite{abadi_tensorflow_2016}. Each RNN consisted of a trainable embedding layer with \textit{embed\_dim} neurons (without pre-trained embeddings), a unidirectional recurrent layer with \textit{hidden\_dim} neurons, and a final dense layer for output. Tokenization was performed using Tensorflow's \textit{TextVectorization} layer. Three types of recurrent cells were evaluated: standard (vanilla) RNN, LSTM \cite{hochreiter_long_1997}, and GRU \cite{cho_learning_2014}. The models were trained on the SNIPS and ATIS intent detection dataset. The training process optimized the multiclass cross-entropy loss function, used for classification tasks. We used Adam optimizer \cite{kingma_adam_2015} for all experiments. Training was performed with early stopping based on validation accuracy \cite{prechelt_early_1996} (patience=2 epochs) to prevent overfitting. 

It is important to note that our goal was not to achieve state-of-art optimized performance, but rather to obtain a reasonably high-performance (e.g., > 93\% accuracy). This ensures the models are competent and their learned dynamics, which are the focus of our study, are meaningful. To this end, hyperparameters were tuned independently for each dataset. For the SNIPS dataset we used a learning rate of $\eta= 5 \times 10^{-4}$ and a batch size of 32. No additional regularization was applied. For the ATIS dataset, we used a learning rate of $\eta= 2.5 \times 10^{-3}$ which was halved after the second epoch. and a batch size of 16. To manage this dataset's complexity and achieve our target performance, we also applied dropout (rate=0.2) \cite{srivastava2014dropout}, to the recurrent layer. The best performing model for each architecture was selected based on its performance in a validation dataset. The validation subsets were created by randomly sampling 20\% of the training data, to ensure a balanced representation of the target classes. The balanced class distribution of the SNIPS dataset made accuracy an appropriate evaluation metric. For the imbalanced ATIS dataset, we report disaggregated F1-scores to provide a more nuanced analysis. All metrics were calculated on a separate test dataset not used during training or validation. Each training procedure was repeated using 10 different seeds.

\section{Results}\label{sec:results}
\subsection{Intent Detection Low-dimensional Dynamics}\label{subsec: low_dim_dynamics}
\begin{figure}[h]
\centering
\includegraphics[width=11cm]{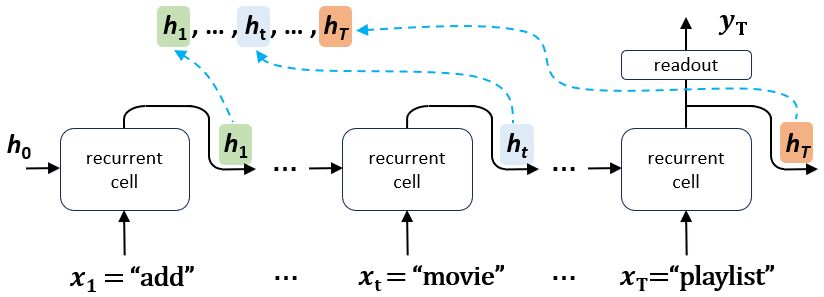}
\caption{Sequence of hidden states $\mathbf{h}_1, \dots, \mathbf{h}_T$ associated with a tokenized input sentence $\mathbf{x}_1, \dots, \mathbf{x}_T$ (e.g. \textit{"add ... movie ... playlist"}) as it is processed by an RNN. The hidden states $\mathbf{h}_t$ evolve in response to input tokens $\mathbf{x}_t$, representing the progression of the network's internal dynamics. The initial hidden state $\mathbf{h}_0$ serves as the starting point before any tokens are processed. The final hidden state $\mathbf{h}_T$ captures the cumulative information of the input sequence and is used to generate the prediction $\mathbf{y}_T$ via the readout layer. The type of recurrent cell (Vanilla, GRU, LSTM) influences how the network captures sequential dependencies.}
\label{fig:hidden_states_inspection}
\end{figure}

In this section, we show that the state space learned by an RNN during the intent detection task is constrained to a low-dimensional hypersurface, or manifold, embedded within the high-dimensional space of the hidden layer. Before sentences are processed by RNNs, a tokenization mechanism transforms each natural language phrase into a sequence $\mathbf{x}_1, \dots, \mathbf{x}_T$, where each token $\mathbf{x}_i \in \mathbb{R}^m$ is an $m$-dimensional vector and $T$ is the number of tokens in sentence \cite{jurafsky_speech_2021}. As shown in Figure~\ref{fig:hidden_states_inspection}, injecting this sequence into an RNN generates a corresponding sequence of activations $\mathbf{h}_1, \dots, \mathbf{h}_T$ in the hidden layer. Each hidden state $\mathbf{h}_i \in \mathbb{R}^n$ is an $n$-dimensional vector (with $n = hidden\_dim$) computed using Equation~\ref{eq_recurrence}. The collection of all hidden states visited by the input sentences constitutes the state space learned by the trained RNN. 

Given that the state space points may lie on a manifold embedded within a higher-dimensional space, the natural question is to determine the intrinsic dimensionality of this manifold. Intrinsic dimensionality refers to the minimum number of dimensions required to accurately represent the variability of the data. Several methods exist to estimate this dimensionality \cite{levina_maximum_2004, camastra_estimating_2002}. Previous studies suggest that the variance-explained threshold is a robust and empirically validated approach, particularly for classification tasks \cite{aitken_geometry_2021}. This method involves performing Principal Component Analysis (PCA) \cite{jolliffe_principal_2002} and determining the number of principal components required to explain a fixed percentage (typically 95\%) of the total variance. 

\begin{figure}[h]
\centering
\includegraphics[width=16.2cm]{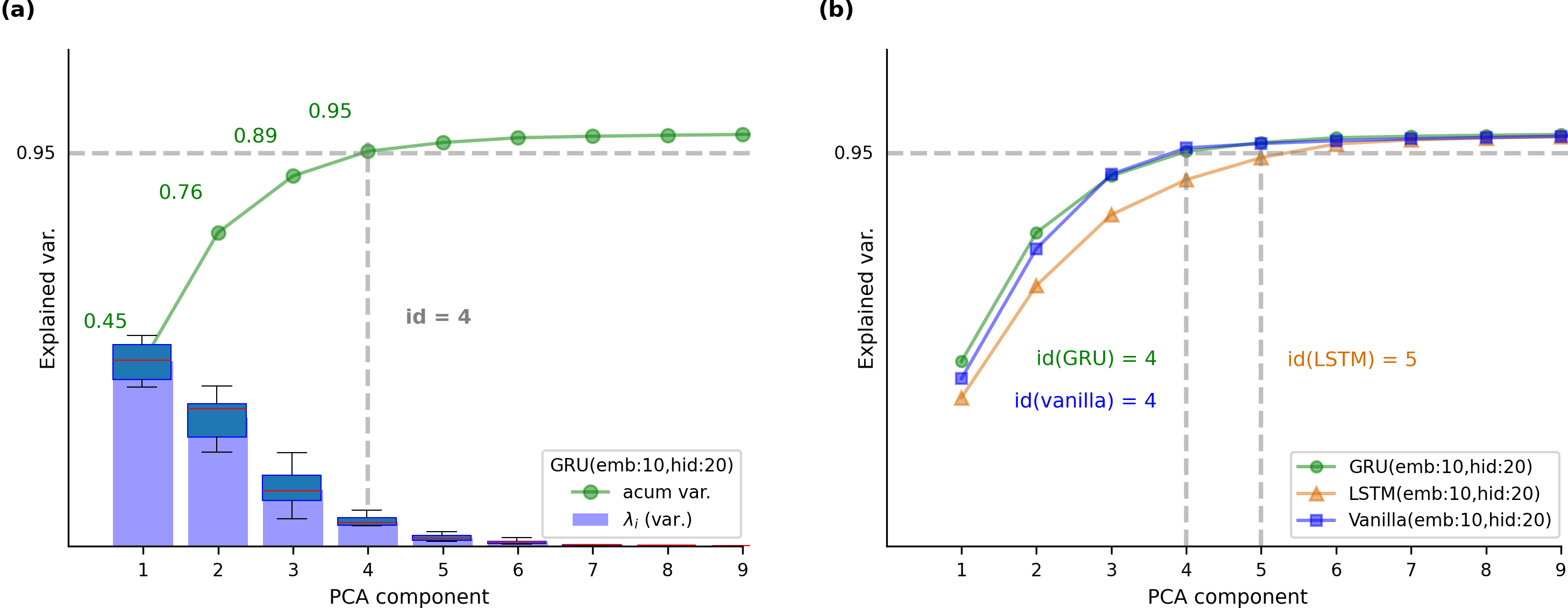}
\caption{Variance explained by principal components for the hidden states of RNNs trained on the SNIPS dataset. \textbf{(a)} PCA analysis for a GRU-based RNN with embedding size = 10 and hidden layer size = 20. The bars represent the variance explained by individual components, while the green curve shows the cumulative variance. The intrinsic dimensionality (id = 4) is marked where the cumulative variance surpasses the 95\% threshold (dashed horizontal line). \textbf{(b)} Comparison of intrinsic dimensionalities for GRU, LSTM, and vanilla RNNs with embedding size = 10 and hidden size = 20. The cumulative variance curves show that GRU and vanilla RNNs reach $id = 4$, while LSTM requires $id = 5$ to exceed the threshold.}
\label{fig:explained_variance}
\end{figure}

Using this approach, we analyzed the state space learned by RNNs trained on the SNIPS dataset. All sentences from the SNIPS test dataset were processed by trained RNNs, and the resulting hidden states were concatenated to form the state space. PCA was then applied to these hidden state points to compute the cumulative variance explained by each principal component. Figure~\ref{fig:explained_variance} illustrates the results for different RNN architectures with $embed\_dim=10$ and $hidden\_dim=20$. The explained cumulative variance is plotted against the number of principal components for the vanilla, LSTM, and GRU cells, represented in blue, yellow, and green, respectively. The variance threshold 95\% is indicated by the horizontal dashed red line, and the number of principal components required to exceed this threshold defines the intrinsic dimensionality $id$ of the state space. As shown in the figure, the GRU and vanilla RNNs reach a $id=4$, while the LSTM networks require a slightly higher dimensionality of $id = 5$.

\begin{figure}[h]
\centering
\includegraphics[width=16.2cm]{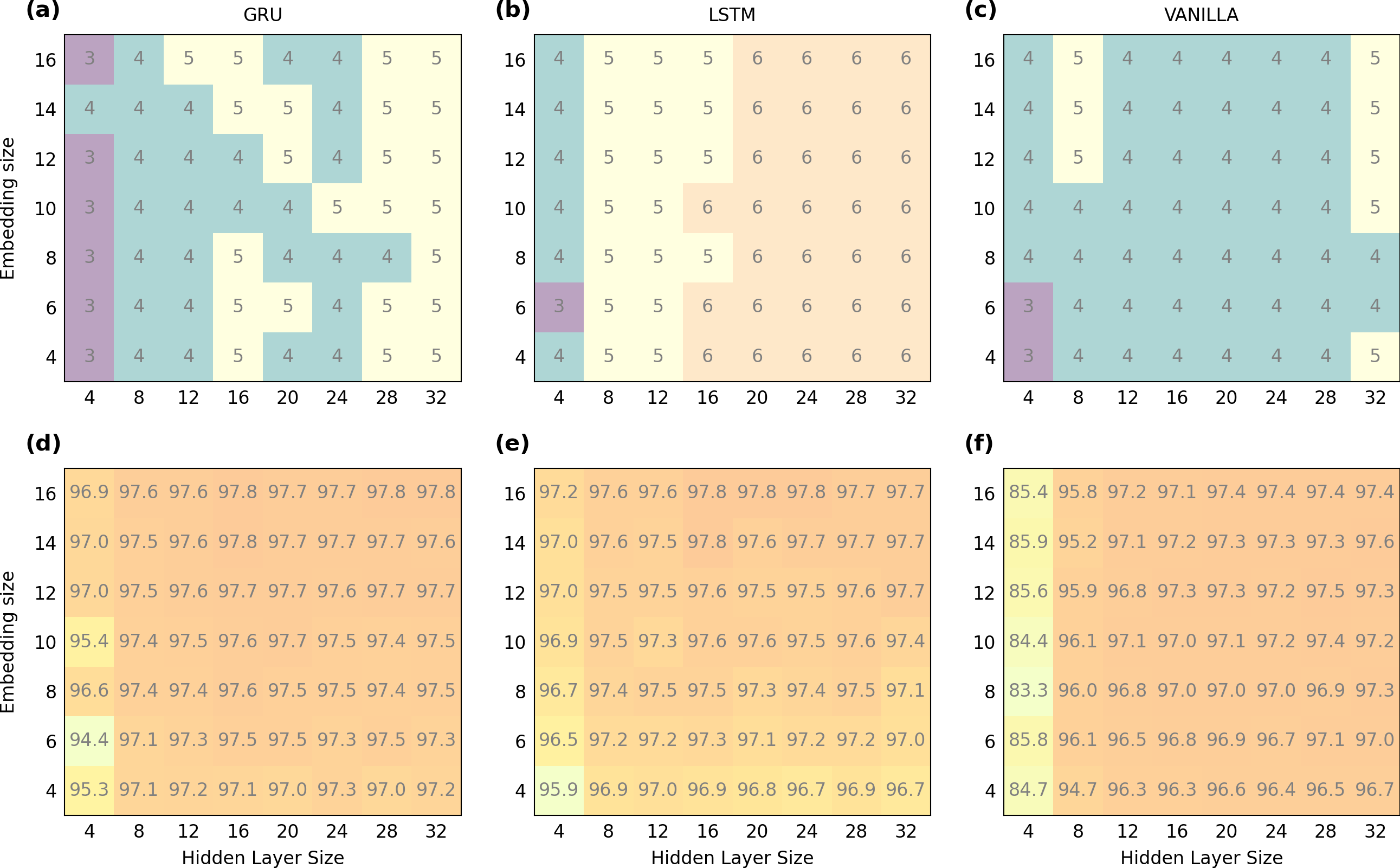}
\caption{State space dimensionality and accuracy of RNNs trained on the SNIPS dataset for different combinations of embedding and hidden layer size. The top row shows the intrinsic dimensionality (id) of the state space for \textbf{(a)} GRU RNNs, \textbf{(b)} LSTM RNNs and \textbf{(c)} Vanilla RNNs. The bottom row displays the corresponding classification accuracy for the same architectures: \textbf{(d)} GRU, \textbf{(e)} LSTM, and \textbf{(f)} Vanilla RNNs.}
\label{fig:dimensionality_greed_search}
\end{figure}

From related work, it is known that the state space dimensionality of RNNs solving categorical text classification tasks is $N-1$, where $N$ is the number of classes \cite{aitken_geometry_2021}. For the SNIPS dataset, which contains 7 intent classes, this prediction implies an expected intrinsic dimensionality $id_e =N-1 = 6$. To investigate whether this hypothesis also holds for intent detection tasks, we performed a comparative analysis of the state space dimensionality and classification accuracy for RNNs with various combinations of embedding layer size and hidden layer size\footnote{For the sake of simplicity, we denote by \textit{cell\_type(emb:x,hid:y)} an RNN with a \textit{cell\_type} recurrent unit, \textit{x} neurons in the embedding layer, and a hidden layer of size \textit{y}.}. Figure~\ref{fig:dimensionality_greed_search} summarizes the results of this analysis. Each point in the figure represents the average classification accuracy and the median state space dimensionality across training runs with 10 different random seeds. Interestingly, the findings reveal that the dimensionality of the state space is not solely determined by the number of classes, $N$, as theoretical predictions might suggest. Instead, it is also significantly influenced by the architectural parameters of the network, including the type of recurrent cell, the size of the embedding layer, and the number of neurons in the hidden layer. In particular, the intrinsic dimensionality ($id$) of the state space is often lower than the theoretically expected dimensionality ($id \leq id_e=N-1$), demonstrating that RNNs can encode task-relevant information in a more compact manner than previously assumed for generic classification tasks.

\subsection{Intent Detection State Space Projection}\label{subsec:state_space_projection}
In the following section, we analyze the spatial organization of hidden states in trained RNNs, taking advantage of the low intrinsic dimensionality ($id < n\_hidden$) of their state spaces. Our analysis reveals that the trained network effectively partitions the state space into distinct clusters. Each cluster corresponds to a specific intent, grouping the hidden states of sentences that share the same intent label.

\begin{figure}[h]
\centering
\includegraphics[width=16.2cm]{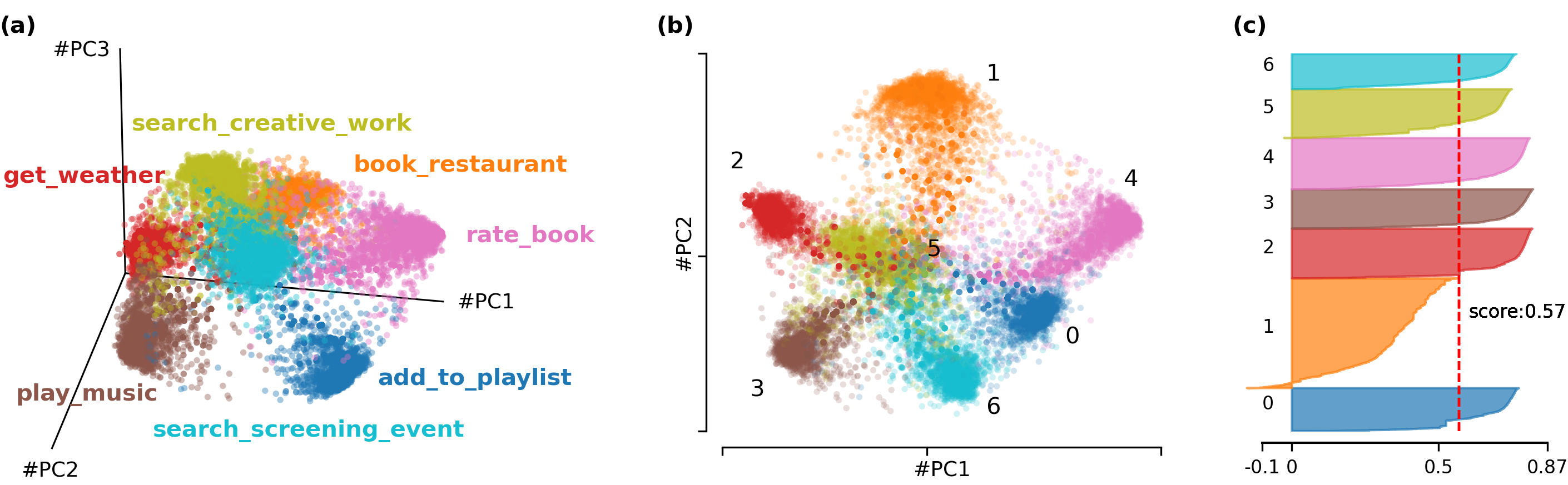}
\caption{Top-2 and top-3 PCA projections of the state space of a GRU(emb:16,hid:16) trained on the SNIPS dataset. The hidden states are colored based on the intent label of its corresponding sentence. \textbf{(a)} 3D PCA projection. \textbf{(b)} 2D PCA projection, highligting intent clusters. \textbf{(c)} Silhouette score from a K-means clustering analysis of the state space, showing a score of 0.57 indicating a moderate level of cluster separation.}
\label{fig:projected_state_space}
\end{figure}

Given a state space, its hidden states can be projected onto a lower-dimensional subspace defined by the top $k$ principal components, as identified in Section~\ref{subsec: low_dim_dynamics}. This projection is achieved by applying a linear transformation $\mathbf{U}$ to each hidden state vector:
\begin{equation}
\hspace{17 em}
\mathbf{p}_i = \mathbf{h}_i\mathbf{U} 
\label{eq:projection}
\end{equation}

\noindent where $\mathbf{h}_i \in \mathbb{R}^n$ is the $n$-dimensional hidden state, $\mathbf{p}_i \in {R}^k$ is the projected $k$-dimensional hidden state, and $\mathbf{U}$ is a $n \times k$ projection matrix whose columns correspond to the top $k$ eigenvectors from the principal component analysis. For visualization purposes, we typically consider projections onto the top-2 and top-3 principal components, which allow the state space to be represented in 2D or 3D, respectively; meanwhile, higher-dimensional projections are useful for dimensionality analysis. Figures~\ref{fig:projected_state_space} (a) and (b) illustrate the 2D and 3D projections, respectively, of the state spaces learned by a GRU(emb:16,hid:16). Each hidden state is colored according to the intent label of the sentence that generated it. The hidden states show a clear clustering pattern, with groups of states corresponding to sentences of the same intent. This clustering suggests that RNNs have successfully learned to organize their state spaces in a way that reflects the semantic distinctions between intents. These visualizations highlight the structured nature of the state space geometry, where each intent is associated with a distinct region in the reduced-dimensional space. The separation between clusters is indicative of the RNNs ability to encode evidence for each intent in an interpretable manner.

To numerically verify the presence of clusters in the state space, we applied the classical K-means clustering algorithm \cite{lloyd_least_1982}, configured with 7 clusters (matching the number of intents in the SNIPS dataset), random initialization of centroids and the Euclidean distance metric. The resulting state space partition was evaluated using the silhouette technique \cite{rousseeuw_silhouettes_1987} which provides a measure of the quality of the clustering. The silhouette coefficient of a point quantifies its separation from other clusters by comparing the average intra-cluster distance (distance to points within the same cluster) to the nearest-cluster distance (distance to points in the closest neighboring cluster). The silhouette coefficients range from [-1, 1]. A value near +1 indicates that the point is well separated from neighboring clusters. Values around 0 indicate points that lie near the decision boundary between two neighboring clusters. Finally, negative values suggest that the point may have been incorrectly assigned to its cluster. The silhouette score is calculated as the mean silhouette coefficient over all points in the dataset. A silhouette score greater than 0.5 is considered indicative of high-quality clusters, where points are well separated and internally cohesive. Figure~\ref{fig:projected_state_space} (c) shows the silhouette scores for a GRU(emb:16,hid:16), where the distances are calculated in the projected state space using only the top-$id$ principal components, with $id$ corresponding to the intrinsic dimensionality of the state space. The silhouette score exceeds the 0.5 threshold, indicating the quality of the clustering and the clear separation of clusters in the reduced-dimensional space. This behavior is consistent across all tested configurations and cell types, as shown by the silhouette scores reported in Table~\ref{tab:last_states_cluster_analysis}, supporting the robustness of the clustering behavior across different architectural variations.

This result confirms the existence of a well-defined partition in the state space, where hidden states corresponding to sentences with the same intent are grouped into distinct regions. Furthermore, this partition can be analyzed without requiring the full-dimensional hidden state space. Instead, the structure of the state space can be captured by examining its intrinsic manifold through the top-$id$ projections. Importantly, the clustering of hidden states into distinct regions based on intent is robust across different architectures and hyperparameter configurations. Regardless of the type of recurrent cell, embedding dimension, or hidden layer size, K-means clustering produces well-defined clusters, reinforcing the idea that intent-specific groupings are an inherent property of the learned state space.

\subsection{Model Inference Mechanism: Sentences as Trajectories}\label{subsec:sentences_trajectories}
In this section, we analyze how input sentences are processed as trajectories through the state space of a trained RNN, revealing structured and predictable behavior within the state space. Sentences evolve along paths, reaching distinct regions of the state space that can be numerically identified as clusters. These regions enable the RNN to classify sentences based on the semantic patterns captured in the final hidden states. 

\begin{figure}[h]
\centering
\includegraphics[width=16.25cm]{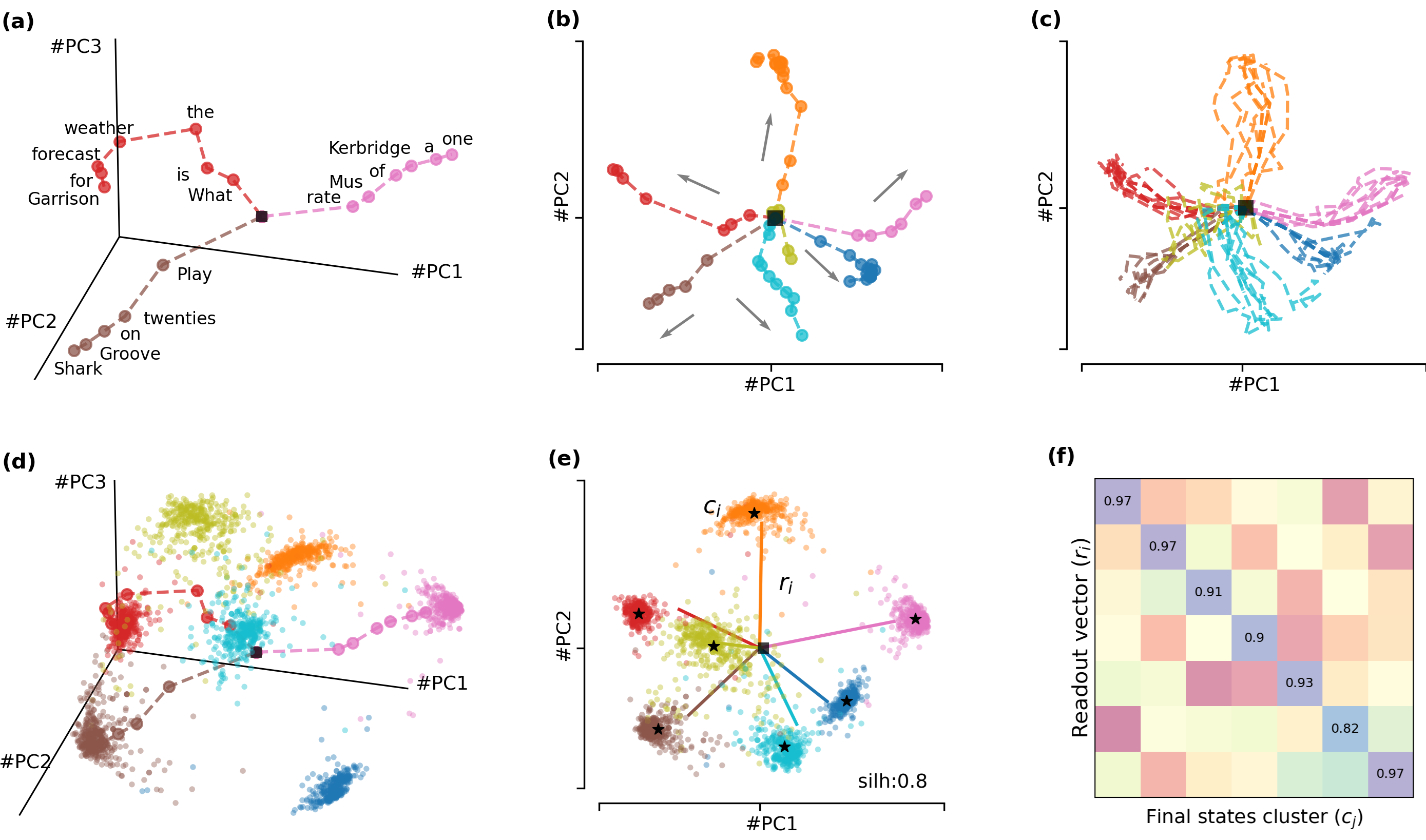}
\caption{State space visualizations of a GRU(emb:16,hid:16) trained on the SNIPS dataset, projected onto the top principal components. \textbf{(a)} Example trajectories for three individual sentences. Each point represents a hidden state corresponding to a token (labeled) in the sentence, a black square marks the initial hidden state. Dashed lines show transitions between states. \textbf{(b)} A representative trajectory for each intent, with arrows indicating the direction of movement. \textbf{(c)} Overlay of multiple trajectories sharing the same intent. \textbf{(d)} Three example trajectories embedded in the 3D state space, superimposed on the final hidden states of all test sentences. \textbf{(e)} Clusters of final states with centroids marked as black stars. Readout vectors ($r_i$) from the output layer are color-coded to match their corresponding clusters ($c_i$). The silhouette score for the clustering is 0.8. \textbf{(f)} Heatmap showing cosine similarities between final state cluster centroids and readout vectors. Only values above 0.5 are displayed.} 
\label{fig:trajectories}
\end{figure}

For each input token $\mathbf{x}_i$, the hidden state $\mathbf{h}_t$ is updated according to Equation~\ref{eq_recurrence}, producing the next state $\mathbf{h}_{t+1}$. As a result, an input sequence $\mathbf{x}_1, \dots, \mathbf{x}_T$ generates a corresponding sequence of hidden states $\mathbf{h}_1, \dots, \mathbf{h}_T$, which describe a trajectory traversing the state space of the RNN. The hidden states can be projected onto the principal components of the $id$-dimensional state space, as described in Equation~\ref{eq:projection}. Figure~\ref{fig:trajectories} (a) illustrates example trajectories for three sentences, each associated with a distinct intent. Each hidden state $\mathbf{h}_i$ is marked with a bullet, while the initial state $\mathbf{h_0}$, representing the zero-value vector before any token is injected, is represented as a black square. Lines connecting the states emphasize the sequential movement through the state space. Specific examples include: red trajectory ("\textit{get\_weather}" intent) "\textit{What is the weather forecast for Garrison}"; brown trajectory ("\textit{play\_music}" intent) "\textit{Play twenties on Groove Shark}" and purple trajectory (\textit{"rate\_book" intent}) "\textit{Rate Mus of Kerbridge a one}". The corresponding input tokens are labeled next to the hidden state they generate. As tokens are processed sequentially, the orbits diverge from the origin and move towards distinct, peripheral regions of the state space. 

This behavior generalizes across all intents and sentences, as shown in Figures~\ref{fig:trajectories} (b) and (c), respectively. Here, individual bullets have been removed for a cleaner representation, and arrows have been added to emphasize the sense of movement along the trajectories. These trajectories appear to direct the hidden states toward specific regions of the state space depending on the intent associated with the sentence. The final hidden state $\mathbf{h}_T$, located at the endpoint of each trajectory, is particularly significant as it determines the prediction of the network. Figure~\ref{fig:trajectories} (d) shows that the endpoints of these trajectories (i.e. the final hidden state) form distinct clusters, each corresponding to a particular intent. To numerically confirm the existence of these clusters, we applied a K-means algorithm (with 7 clusters) to the final hidden states. The results for different RNN configurations are summarized in Table~\ref{tab:last_states_cluster_analysis}. Across all tested configurations and cell types, silhouette scores greater than 0.75 confirm the presence of as many clusters as intents, demonstrating that this clustering behavior is robust and consistent regardless of architectural variations.

As shown in Figure~\ref{fig:hidden_states_inspection}, for prediction purposes, intermediate states are discarded, and only the final hidden state $\mathbf{h}_T$ is considered. Given a sentence, the logits for each class are computed by projecting $\mathbf{h}_T$ onto the readout vectors $\mathbf{r}_i$, corresponding to the rows of the readout matrix $\mathbf{W}$ as follows:
\begin{equation}
\hspace{12 em}
\mathbf{y} = \mathbf{y}_T = \mathbf{W}\mathbf{h}_T = [\mathbf{r}_1|\ldots|\mathbf{r}_n]^T\mathbf{h}_T
\end{equation}

This formulation enables the RNN to classify sentences based on the semantic patterns encoded in the final hidden states. For a sentence with true intent $I$ and final state $\mathbf{h}_T$, correct prediction requires that $\mathbf{r}_I^T\mathbf{h}_T$ be greater than $\mathbf{r}_i^T\mathbf{h}_T$ for any $i \ne I$. To achieve this, the readout vector $\mathbf{r}_I$ must align as closely as possible with the final states $\mathbf{h}_{TI}$ of the cluster associated with intent $I$. Cosine similarity is a widely used measure of vector alignment \cite{dangeti_statistics_2017}, which captures the cosine of the angle between vectors. The values range from -1 (opposite direction) to 1 (perfect alignment), with 0 indicating orthogonality. Figure~\ref{fig:trajectories} (f) shows the cosine similarity between all pairs ($r_i$, $c_j$) for a trained GRU(emb: 16,hid: 16). Each $r_i$ aligns closely with a single centroid, achieving similarity values exceeding 0.9. The visualization in Figure~\ref{fig:trajectories} (e) confirms that the cluster centroids align with the readout vectors of the output layer, as evidenced by the cosine similarity heatmap in Figure~\ref{fig:trajectories} (f). This alignment illustrates how the RNN organizes its state space to facilitate classification, guiding sentence trajectories toward decision regions defined by the output layer. The predicted intent corresponds to the index $i$ with the highest scalar value $\mathbf{r}_i^T\mathbf{h}_T$. In Figure~\ref{fig:trajectories} (e), the final hidden states are projected onto their principal components and colored by intent. The readout vectors $r_i$ are also projected into this space using the same color scheme. 

\subsection{Cluster Spatial Characteristics}
In this section, we analyze the spatial arrangements of the final hidden state clusters in the RNN state space. The results reveal a robust structure across different configurations, with cluster centroids located approximately equidistant from the initial hidden state and clusters remaining compact, enabling effective intent classification. 

\begin{figure}[h]
\centering
\includegraphics[width=16.15cm]{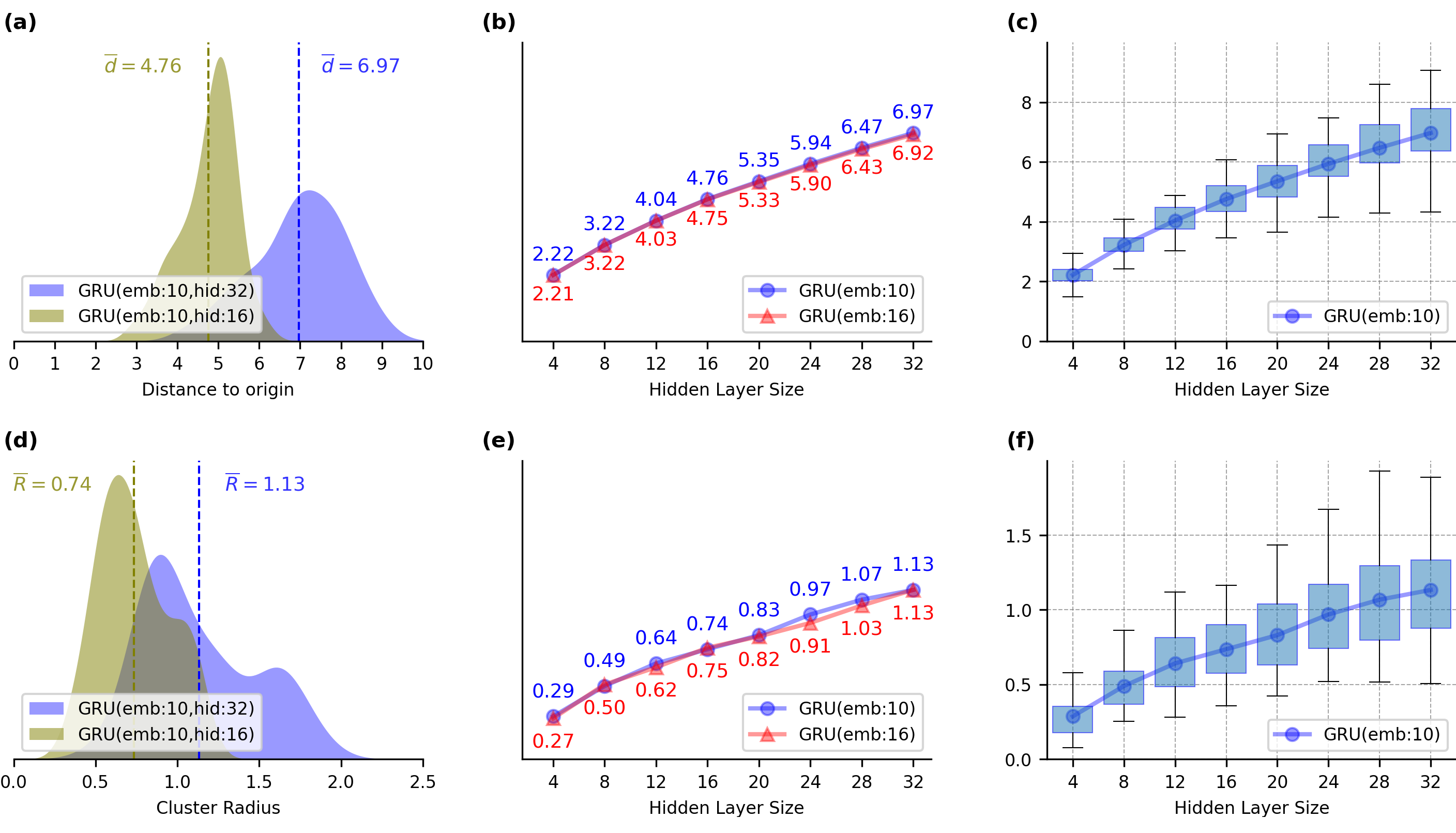}
\caption{Analysis of the spatial characteristics of final hidden states clusters for different GRU configurations. 
\textbf{(a)} Distribution of the distances ($d$) from each cluster centroid to the initial hidden state ($\mathbf{h_0}$), for each configuration. \textbf{(b)} Mean centroid distances $\bar{d}$ as a function of hidden layer size for two different embedding sizes. \textbf{(c)} Variability of centroid distances for GRU with embedding = 10 across hidden layer sizes.\textbf{(d)} Distribution of the clusters radii ($R$) for both configurations. \textbf{(e)} Mean cluster radii ($\bar{R}$) as a function of hidden layer size for two different embedding sizes. \textbf{(f)} Variability of cluster radii for GRU with embedding=10 across hidden layer sizes.}
\label{fig:distances_radii}
\end{figure}

For each network configuration, we calculated the Euclidean distances $d_i$ between the centroid of the $i$-th cluster and the initial hidden state $\mathbf{h_0}$. The distribution of these distances for two different GRU configurations is presented in Figure~\ref{fig:distances_radii} (a). In all configurations, the standard deviation $\sigma(d)$ of these distances is significantly smaller than their mean $\bar{d}$, indicating low variability. This consistency suggests that cluster centroids are roughly equidistant from the initial state. Figure~\ref{fig:distances_radii} (b) shows that the mean centroid distance $\bar{d}$ increases as the size of the hidden layer grows, while remaining largely unaffected by changes in embedding size. This trend reflects the network's ability to distribute information more broadly in the state space as its hidden layer capacity increases. The boxplot in Figure~\ref{fig:distances_radii} (c) further highlights the low variability in centroid distances between different hidden layer sizes.

In addition, we measured the cluster radii $R_i$ obtained as the average Euclidean distance between all points in the $i$-th cluster and its associated centroid $c_i$. The distribution of these radii for two distinct GRU configurations is shown in Figure~\ref{fig:distances_radii} (d). Like centroid distances, the cluster radii increase with the growth of the hidden layer size, as illustrated in Figure~\ref{fig:distances_radii} (e). As centroids move from the origin with larger hidden layers, the clusters expand proportionally. However, compared to centroid distances, the cluster radii presents a greater variability, as shown in Figure~\ref{fig:distances_radii} (f). This greater dispersion reflects inherent differences in how hidden states from different intents are distributed within each cluster. As shown in Table 2, this behavior of distances and radii is consistent across a variety of configurations and cell types, further validating the robustness of the observed patterns in different network architectures.

\begin{table}[h]
\caption{Cluster analysis of final hidden states for various RNN configurations. Silhouette scores for state space partitioning and final states clustering, evaluated in both the original high-dimensional state space and the intrinsic low-dimensional ($id$-projected) space. Statistics for cluster centroid distances ($\bar{d}$, $\sigma(d)$) and cluster radii ($\bar{R}$, $\sigma(R)$). Mean cosine similarity between readout vectors and cluster centroids (readout-cluster alignment).}
\centering 
  \setlength{\tabcolsep}{4.0pt}
  \begin{tabular}{c c c c c c c c c c c c}
    \toprule
      RNN & Embedding & Hidden & 
      \multicolumn{2}{c}{\CellWithForceBreak{Space Partition (Silh.)}} &
      \multicolumn{2}{c}{\CellWithForceBreak{Final Clusters (Silh.)}} &
      \multicolumn{2}{c}{\CellWithForceBreak{Cluster Distances}} &
      \multicolumn{2}{c}{\CellWithForceBreak{Cluster Radii}} &
      Readout-Cluster \\
      cell type & dimension & dimension & Original & $id$-projected & Original & $id$-projected &
      $\bar{d}$ & $\sigma(d)$ & $\bar{R}$ & $\sigma(R)$ & alignment\\
    \midrule
    GRU	    &10	&12	&0.52	&0.53	&0.74	&0.78	&4.04	&0.50	&0.64	&0.20	&0.95 \\
    GRU	    &16	&16	&0.51	&0.52	&0.77	&0.79	&4.75	&0.56	&0.75	&0.21	&0.95 \\
    GRU	    &16	&20	&0.51	&0.54	&0.77	&0.79	&5.33	&0.72	&0.82	&0.24	&0.94 \\
    LSTM	&10	&12	&0.53	&0.54	&0.80	&0.81	&4.35	&0.45	&0.64	&0.26	&0.93 \\
    LSTM	&16	&16	&0.52	&0.54	&0.81	&0.83	&5.08	&0.54	&0.69	&0.26	&0.96 \\
    LSTM	&16	&20	&0.51	&0.52	&0.79	&0.82	&5.72	&0.60	&0.86	&0.33	&0.95 \\
    Vanilla	&10	&12	&0.49	&0.51	&0.72	&0.74	&4.14	&0.48	&0.79	&0.25	&0.94 \\
    Vanilla	&16	&16	&0.50	&0.52	&0.72	&0.75	&4.83	&0.56	&0.91	&0.29	&0.94 \\
    Vanilla	&16	&20	&0.49	&0.52	&0.72	&0.76	&5.46	&0.49	&1.05	&0.31	&0.95 \\
    \bottomrule
  \end{tabular}
\label{tab:last_states_cluster_analysis}
\end{table}

\subsection{Fixed Point Structure}
In the previous section, we show how sentences traverse a low-dimensional state space, guided by the RNN toward final state clusters. This structured behavior suggests the presence of an underlying fixed point topology in the network dynamics. To investigate this, we use the first-order approximation of the RNN dynamics \cite{khalil_nonlinear_2013} around an expansion point $(\mathbf{h}_e, \mathbf{x}_e)$ expressed as:
\begin{equation}
\hspace{10 em}
\mathbf{h}_t \approx \mathbf{F}(\mathbf{h}_e,\mathbf{x}_e) + 
\mathbf{J}^{rec}\mathbf{F}|_{(\mathbf{h}_e,\mathbf{x}_e)} \Delta \mathbf{h}_{t-1} + \mathbf{J}^{inp}\mathbf{F}|_{(\mathbf{h}_e,\mathbf{x}_e)} \Delta \mathbf{x}_t 
\label{eq:expansion_rnn}
\end{equation}
where $\Delta \mathbf{h}_{t-1} =  \mathbf{h}_{t-1} - \mathbf{h}_e$, $\Delta \mathbf{x}_t =  \mathbf{x}_t - \mathbf{x}_e$ and \{$\mathbf{J}^{rec}\mathbf{F}, \mathbf{J}^{inp}\mathbf{F}\}$ are the Jacobian matrices of the update function $\mathbf{F}$ at the expansion point. Specifically, the \textit{recurrent Jacobian} $\mathbf{J}^{rec}\mathbf{F}$ captures the local dynamics of the recurrent structure, and the \textit{input Jacobian} $\mathbf{J}^{inp}\mathbf{F}$ quantifies the sensitivity of the system to input tokens. The elements of these matrices are defined as: 
\begin{equation}
    \hspace{10 em}
    J_{ij}^{rec}\mathbf{F} = \frac{\partial{F_i}}{\partial{h_j}}\Bigr|_{(\mathbf{h}_e,\mathbf{x}_e)} \qquad J_{ij}^{inp}\mathbf{F} = \frac{\partial{F_i}}{\partial{x_j}}\Bigr|_{(\mathbf{h}_e,\mathbf{x}_e)}
\end{equation}
When the expansion point corresponds to a fixed point $\mathbf{h}^*$, such that $\mathbf{h}^* = \mathbf{F}(\mathbf{h}^*,\mathbf{x})$, Equation~\ref{eq:expansion_rnn} reduces to a linear dynamical system:
\begin{equation}
\hspace{8 em}
\Delta \mathbf{h}_t  = \mathbf{h}_t - \mathbf{h}_e \approx
\mathbf{J}^{rec}\mathbf{F}|_{(\mathbf{h}^*,\mathbf{x}^*)} \Delta \mathbf{h}_{t-1} + \mathbf{J}^{inp}\mathbf{F}|_{(\mathbf{h}^*,\mathbf{x}^*)} \Delta \mathbf{x}_t 
\label{eq:expansion_rnn_simplified}
\end{equation}
\noindent Due to time-dependent inputs $\mathbf{x_t}$, RNNs are inherently non-autonomous dynamical systems. This nonautonomy requires advanced mathematical tools for analysis. A reverse engineering approach simplifies the analysis by splitting it into three steps: a) identify the topological structure of fixed points under constant input $\mathbf{x}_t$, b) analyze the linearized system dynamics around these fixed points, and c) examine how nonconstant inputs influence and alter (a.k.a. deflect) the system behavior. \cite{aitken_geometry_2021}.

In this section, we focus on the first step: identifying the fixed point structure. Fixed points $\mathbf{h}_i^*$ are defined as states satisfying $\mathbf{h}_i^* = \mathbf{F}(\mathbf{h}_i^*, \mathbf{x})$ where $\mathbf{x}$ is a constant input. Typically, $\mathbf{x}$ is set to $\mathbf{0}$, representing the system's autonomous evolution without external inputs. In related work, very slow motion points \cite{sussillo_opening_2013} are considered as approximate fixed points. Here, we adopt this broader definition and define fixed points such that $\mathbf{h}_i^* \approx \mathbf{F}(\mathbf{h}_i^*,\mathbf{0})$. Different procedures have been proposed to identify fixed points \cite{katz_using_2018, sussillo_opening_2013}. To numerically identify these points, we minimize the speed $q$ of a point in the state space, defined as the squared magnitude of the displacement generated by $\mathbf{F}$.
\begin{equation}
\hspace{14 em}
q = \frac{1}{2} \|\mathbf{h}-\mathbf{F}(\mathbf{h},\mathbf{0})\|_2^2    
\end{equation}
A numerical optimization process identifies slow-motion ($q < 10^{-8}$) and zero-motion points \cite{golub_fixedpointfinder_2018}. To account for different regions of the state space, this procedure is run with multiple initial conditions. In our experiments, over 25K initial states were extracted from trajectories of the SNIPS test dataset. Under the assumption of $\mathbf{x} = \mathbf{0}$, Equation~\ref{eq:expansion_rnn_simplified} simplifies to:
\begin{equation}
\hspace{12 em}
\Delta \mathbf{h}_t  = \mathbf{h}_t - \mathbf{h}_e \approx
\mathbf{J}^{rec}\mathbf{F}|_{(\mathbf{h}^*,\mathbf{0})} \Delta \mathbf{h}_{t-1}
\label{eq:expansion_rnn_simplified_x0}
\end{equation}
The stability of each fixed point $\mathbf{h^*_i}$ is determined by the eigenvalues of  $\mathbf{J}^{rec}\mathbf{F}|_{(\mathbf{h}_i^*,\mathbf{0})}$. Table \ref{tab:fps_facts} presents the fixed points identified for various RNN configurations trained on the 7-class SNIPS dataset, both stable and saddle fixed points are identified. The number of critical points depends on the type of recurrent cell as well as the dimensions of the embedding and hidden layers. The presence of saddle points with indices higher than one indicates a hierarchy of critical points that trajectories traverse during sentence processing. We further analyze the spatial arrangement of these fixed points by projecting them onto the $id$-top principal components of the state space. The distances $\delta_s$, $\delta_1$, and $\delta_2$ (corresponding to stable points, 1-index saddles and 2-index saddles, respectively) were calculated from each fixed point to the origin. Table~\ref{tab:fps_facts} summarizes the mean $\bar{\delta}$ and standard deviation $\sigma(\delta)$ of these distances for different network configurations. The distances vary with the embedding and hidden layer dimensions, reflecting how architectural parameters influence the state space geometry. 

\begin{table}[h]
\caption{Summary of approximated fixed points ($q < 10^{-8}$) identified in RNNs trained on the SNIPS dataset. The table list the number of attractors, 1-index saddle points, and higher-index saddle points for different network configurations. The mean ($\bar{\delta}$) and standard deviation ($\sigma(\delta)$) of distances from the origin to attractors ($\delta_s$), 1-index saddle points ($\delta_1$) and 2-index saddles ($\delta_2$) are presented for each configuration.}
\centering
\begin{tabular}{cccccccccccc} 
\toprule
    RNN &
    Embedding &
    Hidden &
    Stable &
    \multicolumn{2}{c}{Saddle points} &
    \multicolumn{2}{c}{Stable points}  &
    \multicolumn{2}{c}{1-index}  &
    \multicolumn{2}{c}{2-index}  \\
    
    cell type & dimension & dimension & points & 1-index & Higher-index &
    $\bar{\delta}_s$ & $\sigma(\delta_s)$ & $\bar{\delta}_1$ & 
    $\sigma(\delta_1)$ & $\bar{\delta}_2$ & $\sigma(\delta_2)$\\
\midrule
Vanilla & 10 & 10 & 5 & 9 & 7 & 3.93 & 0.10 & 3.52 & 0.13 & 3.07 & 0.31 \\
Vanilla & 16 & 16 & 4 & 5 & 4 & 4.73 & 0.29 & 4.25 & 0.38 & 4.03 & 0.36 \\
GRU     & 10 & 10 & 5 & 6 & 1 & 3.72 & 0.32 & 3.43 & 0.33 & 1.72 & - \\
GRU     & 16 & 16 & 3 & 3 & 1 & 4.71 & 0.15 & 4.27 & 0.18 & 3.29 & - \\
\bottomrule
\end{tabular}
\label{tab:fps_facts}
\end{table}

\subsection{Generalizability and Explanatory Power: A Case Study on the ATIS Dataset}\label{subsec:atis_case_study}
To address the generalizability of findings beyond the balanced SNIPS dataset, we replicated our analysis on the ATIS dataset. As detailed in section~\ref{sec:datasets}, ATIS presents a more realistic and challenging scenario characterized by a severe class imbalance, with 26 intents (vs. 7 in SNIPS) and the top intent (\textit{flight}) accounting for 73,7\% of samples. This allows us to test if our dynamical systems framework can provide insights under less-controlled conditions. 

We trained a GRU(emb:64,hid:64) architecture on the imbalanced ATIS dataset. To clarify the scope of the analysis: the full ATIS dataset contains 26 intents, but the standard training set contains only 22 of these. The 4 intents that appear only in the test dataset were excluded from our analysis as they represent an out-of-scope task. Furthermore, 6 of the 22 training intents have no samples in the provided test dataset. Our analysis, therefore, focuses on the 16 intents (shown in Table~\ref{tab:atis_scores}) for which samples were present in both the training and the test data. On this 16-class task, the model achieved a high aggregate accuracy (93.7\%), this single metric obscures severe performance disparities. For instance, as shown in Table~\ref{tab:atis_scores}, high-frequency intents like \textit{flight} achieve an F1-score of 0.97, while low-frequency intents like \textit{meal} (6 training samples) fail completely with a 0.00 F1-score. Standard metrics show that the model fails on rare classes, but our framework can help explain why by analyzing the learned state space.

\begin{figure}[h]
\centering
\includegraphics[width=16.25cm]{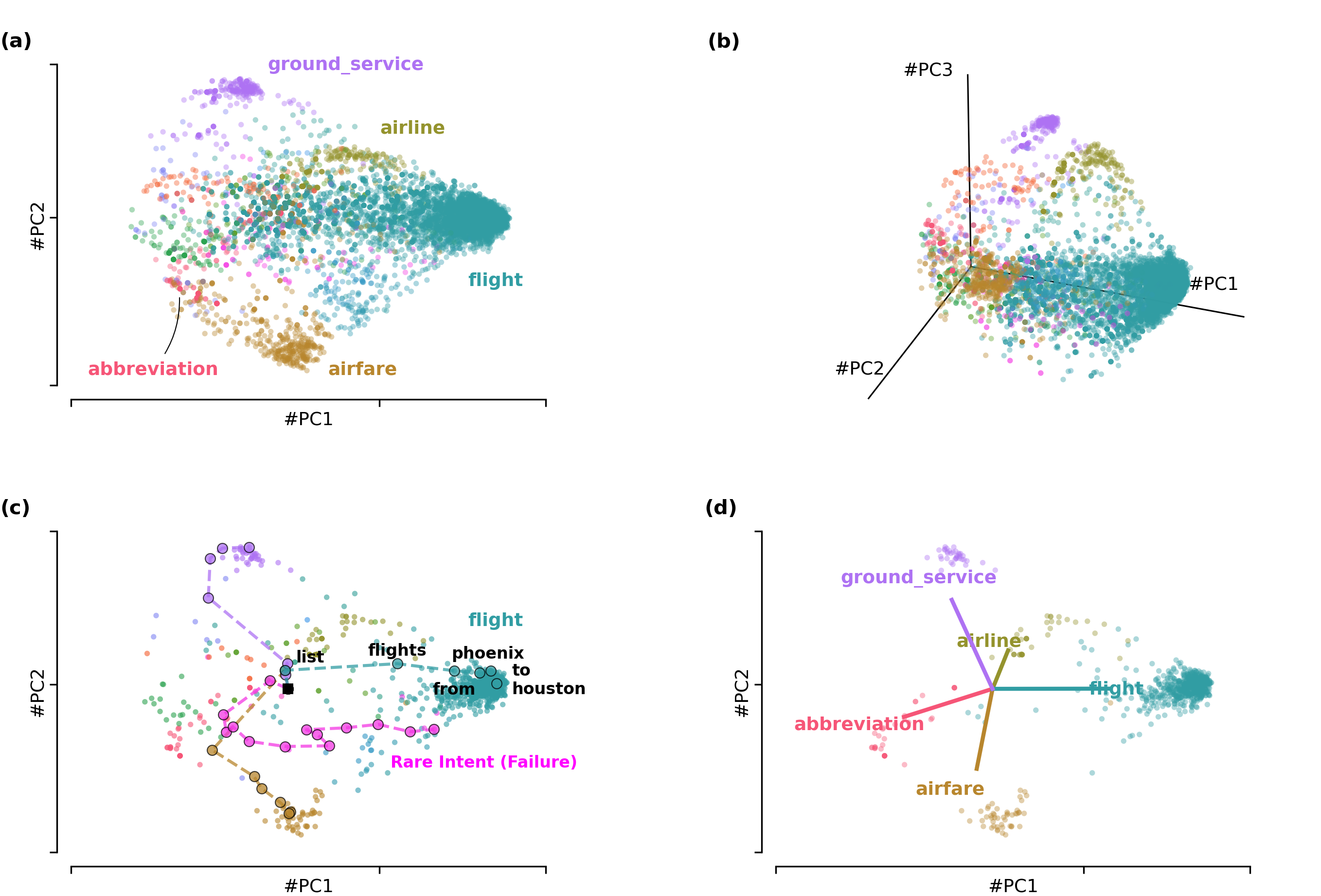}
\caption{State space visualizations of the GRU model trained on the imbalanced ATIS dataset. \textbf{(a)} 2D projection of the final hidden states, showing a few large distinct clusters for high-frequency intents (e.g. flight, airfare). \textbf{(b)} 3D projection, confirming the geometric separation of the major clusters. \textbf{(c)} Example 2D trajectories. A "success" (teal) steers to the flight cluster, while a "failure" (magenta) for a rare intent wanders into the wrong cluster. \textbf{(d)} Alignment of readout vectors (colored lines) with their corresponding high-frequency clusters (colored dots)} 
\label{fig:atis_plots}
\end{figure}

Consistent with our findings on SNIPS, the computation of a GRU(emb:64,hid:64) on ATIS operates on a low-dimensional manifold. A PCA analysis of the test hidden states revealed that only 9 components are required to explain 95\% of the variance, a dimensionality far lower than the model's actual hidden and embedding dimension (64) and the number of classes (16). However, the geometry of this manifold is significantly influenced by the class imbalance. As visualized in the 2D and 3D projections in Figure~\ref{fig:atis_plots} (a) and (b), the state space is dominated by large, well-separated clusters for high-frequency intents (e.g. \textit{flight}, \textit{airfare}). In contrast, most low-frequency intents do not form distinct clusters but instead coalesce into a dense, undifferentiated region. 

This "geometric collapse" of rare classes makes standard clustering algorithms like K-means ill-suited for this analysis. K-means, which seeks to find well-defined centers, would fail to partition this dense region and would misrepresent the landscape. Therefore, to properly quantify the quality of the partition the network was supposed to learn, we used a methodology based on the ground-truth labels. We posit that this geometric structure provides a powerful lens to diagnosing classification performance. We use this lens to test our hypothesis that successful classification requires the network to accomplish two distinct tasks, which we evaluated consistently within the 95\% variance projected space:
\begin{itemize}
    \item \textbf{Geometric Separation:} The network must guide trajectories for a given intent to a unique and coherent region of the state space. This dynamic steering process is visualized in Figure~\ref{fig:atis_plots} (c), which contrasts a successful \textit{flight} trajectory with a failed one for a rare intent. 
    \item \textbf{Readout Alignment:} The network must correctly align the corresponding readout vector $(r_i)$ with that specific geometric region $(c_i)$.  This static link between the final region and the output layer is visualized for the high-frequency intents in Figure~\ref{fig:atis_plots} (d).
\end{itemize}

\begin{table}[h]
\caption{Per-intent performance and state space diagnostics for a GRU(emb:64,hid:64) model trained on the ATIS dataset. F1-score and the two metrics from our framework are shown: Silhouette Score (Geometric Separation) and Cosine Similarity (Readout Alignment), computed within the 95\% PCA projected state space.}
\begin{tabular}{l r r  ccc cc}
\multirow{2}{*}{\textbf{Intent Class}} & \multirow{2}{*}{\textbf{\makecell{Test\\ \#}}} & \multirow{2}{*}{\textbf{\makecell{Train\\ \#}}} & \multicolumn{3}{c}{\textbf{Performance}} & \multicolumn{2}{c}{\textbf{Projected state space (@95\%)}}\\

\cmidrule(lr){4-6} \cmidrule(lr){7-8} 
& & & \textbf{Precision} & \textbf{Recall} & \textbf{F1-score} & \textbf{Silh. Score} & \textbf{Cosine Sim.} \\
\midrule                                                    
flight	        &632 &3666 &0.96	&0.99	& 0.97	 &0.57     &0.95\\
airfare	        &48  &423  &0.94	&0.98	& 0.96	 &0.61     &0.89\\
ground\_service	&36  &255  &0.84	&1.00	& 0.91	 &0.76     &0.88\\
airline	        &38  &157  &0.90	&0.92	& 0.91	 &0.62     &0.92\\ 
abbreviation	&33  &147  &0.97	&1.00	& 0.99	 &0.50     &0.95\\
flight\_time	&1   &54   &1.00	&1.00	& 1.00   &0.00     &0.90\\
airport	        &18  &20   &0.93	&0.78	& 0.85	 &0.14     &0.94\\
capacity	    &21  &16   &1.00	&0.76	& 0.86	 &0.31     &0.80\\
aircraft	    &9   &81   &0.75	&0.67	& 0.71	 &-0.13    &0.91\\
ground\_fare	&7   &18   &0.50	&0.57	& 0.53	 &-0.07    &0.80\\
quantity	    &3   &51   &0.38	&1.00	& 0.55	 &0.77     &0.88\\
flight\_no	    &8   &12   &1.00	&0.25	& 0.40	 &0.56     &0.57\\
flight+airfare	&12  &21   &0.83	&0.42	& 0.56	 &0.03     &0.66\\
city	        &6   &19   &0.67	&0.33	& 0.44	 &0.02     &0.24\\
distance        &10  &20   &1.00	&0.20	& 0.33	 &-0.20    &0.42\\
meal	        &6   &6    &0.00	&0.00	& 0.00   &-0.21    &0.38\\
\bottomrule
\end{tabular}
\label{tab:atis_scores}
\end{table}

\begin{figure}[h]
\centering
\includegraphics[width=16.25cm]{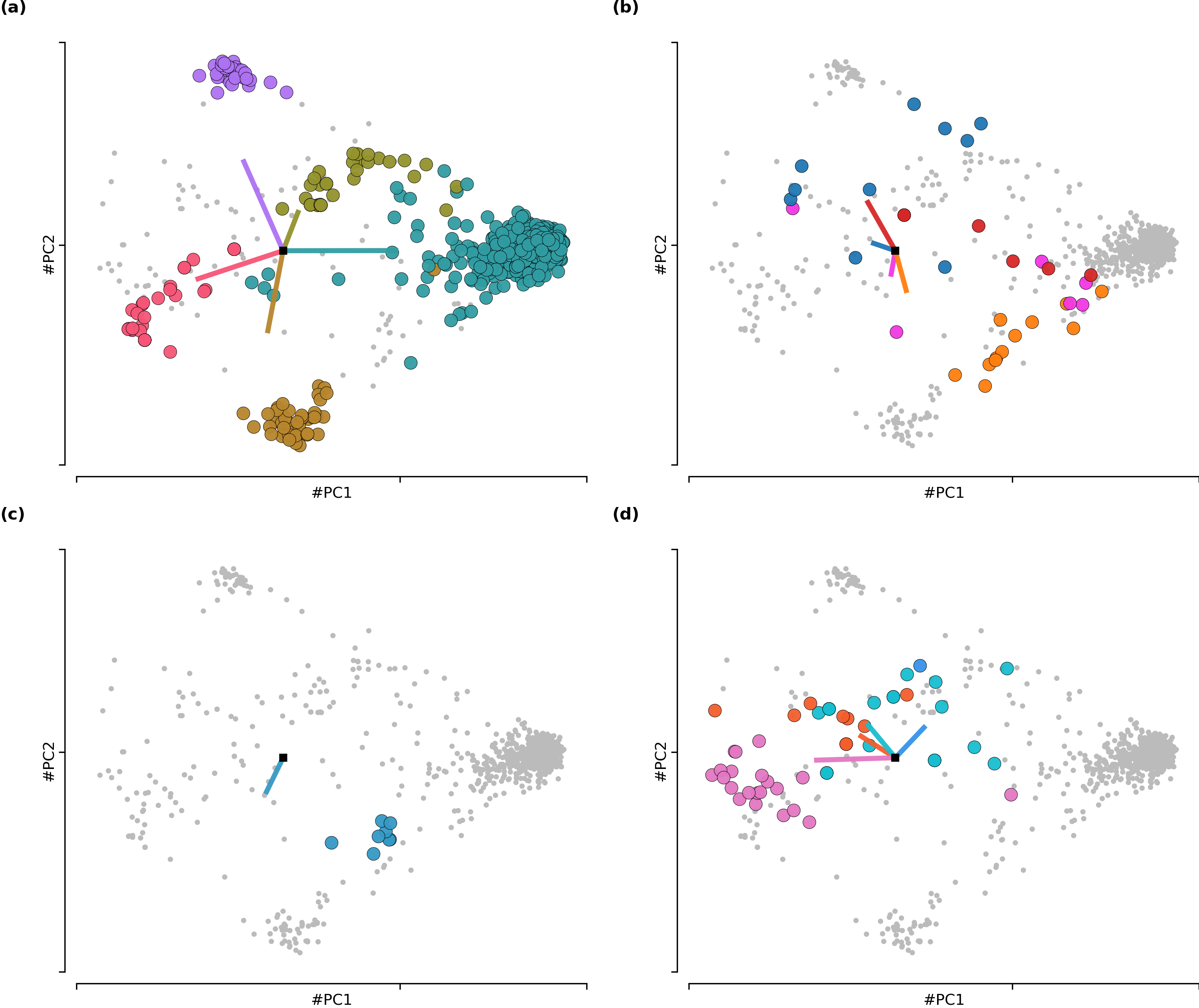}
\caption{The four patterns of model behavior in the imbalanced ATIS PCA state space. Final states for analyzed intents (colored dots), all other final states (Gray dots) and readout vectors for analyzed intents (colored lines). \textbf{(a)} Pattern 1: Convergent High Performance. High separation and strong alignment. \textbf{(b)} Pattern 2: Geometric Collapse. Poor separation and weak alignment. \textbf{(c)} Pattern 3: Alignment Failure. Good separation but poor alignment. \textbf{(d)} Pattern 4: Alignment-Driven Classification. Poor separation but strong alignment.} 
\label{fig:atis_patterns}
\end{figure}

A failure in either step can lead to a poor F1-score. Analyzing these two metrics, as shown in Table~\ref{tab:atis_scores}, reveals distinct, interpretable patterns of model behavior that are invisible to a standard F1-score analysis. These four patterns are visualized in Figure~\ref{fig:atis_patterns}:
\begin{itemize}
    \item \textbf{Pattern 1: Convergent High Performance} (High F1, High Separation, High Alignment). These intents (e.g.\textit{ flight}, \textit{airfare}, \textit{airline}, \textit{ground\_service}, \textit{abbreviation}) represent robust learning. They have high F1-scores, a direct consequence of success in both steps. This ideal case is visualized in Figure~\ref{fig:atis_patterns} (a), showing distinct clusters (with high Silhouette scores) and strongly aligned readout vectors.
    
    \item \textbf{Pattern 2: Geometric Collapse} (Low F1, Low Separation, Low Alignment). These intents (e.g. \textit{meal}, \textit{distance}, \textit{city}, \textit{flight+airfare}) represent a total mechanistic failure. Their near-zero F1-scores are correlated with a failure in both metrics. As shown in Figure~\ref{fig:atis_patterns} (b), these intents fail to form a coherent cluster (negative or near-zero Silhouette scores) and collapse into a dense, mixed region. This is accompanied by a failure in the second step, as shown by their very low readout alignment scores.
    
    \item \textbf{Pattern 3: Alignment Failure} (Low F1, High Separation, Low Alignment). This pattern is clearly exemplified by \textit{flight\_no}. Figure~\ref{fig:atis_patterns} (c) illustrates this pattern perfectly: an intent forms a reasonably distinct cluster (Silhouette 0.56), yet has a very low F1-score (0.40). Table~\ref{tab:atis_scores} and the visualization both identify the cause: a poor readout alignment (0.57). The network successfully grouped this intent, but failed to map the correct output neuron to that group.
    
    \item \textbf{Pattern 4: Alignment-Driven Classification} (High F1, Low Separation, High Alignment). This is the inverse finding, seen in intents like \textit{aircraft}, \textit{airport}, \textit{capacity} and \textit{flight\_time}). These intents achieve high F1-scores (ranging from 0.71 to 1.00) despite exhibiting very poor cluster separation. Figure~\ref{fig:atis_patterns} (d) provides the visual description for this mechanism. A negative score, as in \textit{aircraft}, confirms its samples are, in average, geometrically closer to a neighboring cluster. However, the high F1-score is correlated with the second metric: high readout alignments. This high alignment score means the correct readout vector is strongly aligned with the centroid of its true samples. Therefore, the classification succeeds not because the clusters are geometrically separate (which the Silhouette score proves they are not), but because the readout vector is so precisely aimed at the center of its correct (but messy) cluster that it still "claims" its samples. The projection of these samples onto the correct readout vector is still higher than their projection onto any other vector, resulting in a correct classification. This shows a case where the network's alignment mechanism can succeed despite a failure to create geometrically pure clusters. 
\end{itemize}

Other low-frequency intents, such as \textit{quantity} and \textit{ground\_fare}, show other combinations of metric failures. While these represent complex failure modes, the four primary patterns identified and visualized cover the most distinct and interpretable modes of model success and failure.

This case study on ATIS shows the robustness and explanatory power of our framework. It reveals how dataset properties like class imbalance are not abstract statistical issues but are directly encoded into the geometric and topological structure of the network's learned dynamics.

\section{Conclusion}
Intent detection remains a challenging problem that has yet to be fully solved. Conceptually, one can think of intent as evolving dynamically: as more words are processed, the potential intent becomes more constrained, as if moving between interpretations. Our approach embraces this notion, modeling the search for the final intent as a dynamical process within the state space of a Recurrent Neural Network. 

In this paper, we applied reverse engineering techniques to study the computational mechanisms of RNNs trained for intent detection. Our analysis of the balanced SNIPS dataset revealed that networks converge to an elegant and highly interpretable solution: they partition their state space into a low-dimensional manifold of distinct, well-separated clusters corresponding to each intent. We showed that sentences evolve along structured trajectories, steering the network's hidden state toward the correct cluster. To test the generalizability of this framework, we extended our analysis to the imbalanced ATIS dataset. This more challenging, real-world scenario revealed how this ideal geometric solution is distorted by class imbalance. Our two-part diagnostic framework (Geometric Separation and Readout Alignment) allowed us to move beyond simple accuracy scores and identify four distinct, mechanistic patterns of model success and failure, explaining why certain intents perform well while others fail.

While other interpretability methods can identify which input token are salient, our dynamical systems approach provides a unique, mechanistic understanding of how the network's internal state evolves over time and arrive at a decision. This perspective opens several promising avenues for future research. This geometric framework can be extended to address related, high-stakes problems in conversational AI. Our finding in the ATIS dataset provides a concrete empirical basis for this. For example, in out-of-scope (OOS) detection, utterances would likely produce trajectories that fail to converge to any of the established, high-frequency clusters, terminating instead in a dense, undifferentiated central region. This geometric distance from a final state to the nearest cluster could thus serve as a robust signal for OOS detection. The framework could also be adapted to analyze the joint intent detection and slot filling task. This would make it possible to explore how the state space dynamics for intents and slots interact and mutually influence one another. Additionally, the geometric separation of clusters could serve as a new metric for model robustness, or be used to probe for demographic biases in how different user utterances are processed \citep{sanchez-karhunen_bias_2025}. Finally, a critical and exciting line of work involves adapting this dynamical systems framework from RNNs to the modern Transformer architectures. Understanding the state-space geometry and attractors within these more complex models is a key challenge for the field of interpretability.



\nocite{*}
\bibliographystyle{ios1}           
\bibliography{EJAI_2025/aic_template}        

%

\end{document}